\documentclass[sigconf,nonacm,screen]{acmart}
\usepackage{xcolor}
\definecolor{linkblue}{HTML}{175A9E}

\AtBeginDocument{%
  \hypersetup{
    colorlinks=true,
    citecolor=linkblue,
    linkcolor=linkblue,
    urlcolor=linkblue
  }
}
    
\usepackage{graphicx} %
\usepackage{xspace}
\usepackage[table]{xcolor}

\usepackage{tikz}
\usetikzlibrary{arrows.meta}
\usepackage[T1]{fontenc}
\usepackage{multirow}
\usepackage{tabularx}
\usepackage[most]{tcolorbox}
\usepackage[inline]{enumitem}
\usepackage{cleveref}
\usepackage{booktabs}
\usepackage{subcaption}
\usepackage{tcolorbox}
\usepackage[normalem]{ulem}

\newcommand{\xhdr}[1]{{\vspace{1mm}\noindent{\textbf{#1}}}}

\newcommand{\claude}{{\tt Claude}\xspace} %
\newcommand{\gpt}{{\tt GPT-5}\xspace}
\newcommand{\gptoss}{{\tt GPT-OSS}\xspace} %
\newcommand{\qwen}{{\tt Qwen}\xspace} %

\newcommand{\eg}[0]{\textit{e.g.}\xspace}
\newcommand{\ie}[0]{\textit{i.e.}\xspace}
\newcommand{\vs}[0]{\textit{vs.}\xspace}

\definecolor{c1}{cmyk}{0,0.6175,0.8848,0.1490} 
\definecolor{c2}{cmyk}{0.1127,0.6690,0,0.4431} 
\definecolor{c3}{cmyk}{0.3081,0,0.7209,0.3255} 
\definecolor{c4}{cmyk}{0.6765,0.2017,0,0.0667} 
\definecolor{c5}{cmyk}{0,0.8765,0.7099,0.3647} 

\newtcbox{\hlprimarytab}{on line, rounded corners, box align=base, colback=c3!10,colframe=white,size=fbox,arc=3pt, before upper=\strut, top=-2pt, bottom=-4pt, left=-2pt, right=-2pt, boxrule=0pt}
\newtcbox{\hlsecondarytab}{on line, box align=base, colback=red!10,colframe=white,size=fbox,arc=3pt, before upper=\strut, top=-2pt, bottom=-4pt, left=-2pt, right=-2pt, boxrule=0pt}

\newcommand{\dashifted}{\raisebox{0.5\depth}{\tiny$\downarrow$}}
\newcommand{\uashifted}{\raisebox{0.5\depth}{\tiny$\uparrow$}}

\newcommand{\uag}[1]{{\scriptsize\hlprimarytab{\uashifted{#1}}}}
\newcommand{\dab}[1]{{\scriptsize\hlsecondarytab{\dashifted{#1}}}}

\definecolor{jsonbg}{RGB}{247,247,247}
\definecolor{jsonkey}{RGB}{35,80,140}

\lstdefinelanguage{json}{
    basicstyle=\ttfamily\tiny,
    string=[s]{"}{"},
    stringstyle=\color{jsonkey},
    comment=[l]{//},
    commentstyle=\color{gray},
    showstringspaces=false,
    breaklines=true,
    breakatwhitespace=false,
    columns=fullflexible,
    keepspaces=true
}

\lstdefinestyle{jsonbox}{
    language=json,
    backgroundcolor=\color{jsonbg},
    frame=single,
    rulecolor=\color{gray!40},
    framesep=4pt,
    xleftmargin=2pt,
    xrightmargin=2pt,
    aboveskip=3pt,
    belowskip=0pt
}

\setcopyright{none}

\title{Lost in Delegation: Studying the Effectiveness of Task Orchestration in Generative Search}

\author{Tamara Stojanovska, Elisabeth Kirsten, Leon Swazinna, and Muhammad Bilal Zafar}
\affiliation{%
  \institution{Ruhr University Bochum and UA Ruhr Research Center for Trustworthy AI}
  \city{Bochum}
  \country{Germany}
}
\email{[tamara.stojanovska, elisabeth.kirsten, leon.swazinna, bilal.zafar]@rub.de}

\begin{document}

\begin{abstract}
LLMs have quickly evolved from models used in isolation to agentic systems that delegate tasks to \textit{external tools} and  \textit{other LLMs} to solve complex assignments. This agentic augmentation has led to impressive capability gains. However, when and how delegation improves performance remains poorly understood. For example, it is unclear whether delegation helps when tasks involve post-cutoff information that the orchestrating LLM cannot reliably resolve from its own knowledge. In this work, we use purpose-built datasets with pre- and post-cutoff splits to compare various delegation modes. We find that delegation provides no consistent accuracy improvement over non-delegated setups, with effects varying across models and tasks. When tasks involve knowledge unavailable to the orchestrator, delegation can substantially degrade accuracy, while direct tool use with the same LLM performs much better. 
Our analysis identifies distorted delegated queries, inefficient tool use, and failures to preserve and use information between agents as recurring failure modes. These findings suggest that effective orchestration requires both well-calibrated delegation decisions and effective exchange of information between agents.
\end{abstract}

\maketitle

\section{Introduction}

LLMs are changing how users search the Web. For over two decades, web search relied on Bing- and Google-style search engines where a user typed a short, self-contained query, received a ranked list of results, and followed one or more pages to manually assemble the answer. LLM-powered search engines like Google AI Overview and ChatGPT work differently. They  accept long, complex queries, retrieve information from a variety of sources using search tools or multiple \textit{search agents}, and synthesize it into a single coherent answer~\cite{kirsten2026characterizing,xi-etal-2026-survey,li2026aisearchparadigm}. This new mode of search is spreading quickly~\cite{reid_google_ai_2026}. According to a recent Pew survey, 42\% of US adults now use generative AI chatbots to search for information~\cite{pew_ai_survey_2026}. 

Despite its success, key design aspects of this LLM-driven agentic search  remain understudied.
Given a complex task, the LLM handling the search could approach it using vastly different strategies.
Consider, for instance, the search prompt: ``Report the release year of the following books: \textit{To Kill a Mockingbird}, \textit{The Lord of the Rings}, \textit{The Hobbit}, \textit{Katabasis} and \textit{The Book Thief}''. The model could employ \textbf{(i)} a \textit{single-agent strategy}, where it performs a separate \texttt{\small web\_search} for each book and then compiles the final answer, or \textbf{(ii)} a \textit{delegation-based strategy}, where it acts as an orchestrator and delegates each book to a separate LLM agent and combines their answers.
Although both strategies address the same subtasks, they have different \textit{computational} and \textit{performance} implications.

As shown in Figure~\ref{fig:fig1_setup}, the \textit{single-agent strategy} places all queries and \texttt{\small web\_search} tool outputs in a single trajectory, increasing the KV-cache size~\cite{pope_kvcache} and the cost of attention computation~\cite{NIPS2017_3f5ee243}.
Longer contexts can also lead to ``context-rot'', where model performance degrades as context length increases~\citep{hsieh2024ruler,liu-etal-2024-lost,zhang2026recursivelanguagemodels}.
The \textit{delegation-based strategy} instead uses the fact that \textit{each of the five subtasks is independent}, delegating the token-intensive retrieval to separate agents and only gathering their responses into a final answer. 
Crucially, the evidence retrieved for question $i$ therefore does not need to remain in context while answering an independent question $j$.

The potential benefits of delegation are compelling: independent subtasks can be solved in separate, shorter contexts, reducing context growth and potentially mitigating the computational costs and performance degradation associated with long contexts.
These potential benefits motivate our key question:

\begin{tcolorbox}[
    colback=gray!10,
    colframe=black,
    boxrule=0.6pt,
    arc=0pt, 
    left=6pt, 
    right=6pt, 
    top=4pt, 
    bottom=4pt,
]
    \textbf{Does delegating independent search tasks to subagents improve accuracy over a single agent performing the same subtasks sequentially?}
\end{tcolorbox}

\noindent
Although prior work has explored orchestrated web search setups~\citep{anthropic2025multiagent,xi-etal-2026-survey,li2026aisearchparadigm}, 
a thorough analysis of 
how separating orchestrator and executor contexts affects performance is lacking.
This question is central to designing agentic search and deep research systems, which increasingly mediate access to information on the Web.

\definecolor{txt}{HTML}{3A3A42}
\definecolor{frame}{HTML}{9A9AA5}
\definecolor{answerbg}{HTML}{F4F4F6}
\definecolor{arrowc}{HTML}{8A8F99}
\definecolor{agentbg}{HTML}{F1F3F7}
\definecolor{agentln}{HTML}{5A6478}
\definecolor{pagebg}{HTML}{D9D9E0}
\definecolor{pageln}{HTML}{A6A6B0}
\definecolor{yearbg}{HTML}{EDEFF3}
\definecolor{yearln}{HTML}{6E7A8C}
\definecolor{yeartx}{HTML}{46505F}

\definecolor{acc}{HTML}{2760A5}\definecolor{accbg}{HTML}{E5EFFD} %
\definecolor{note}{HTML}{D9731A}\definecolor{notebg}{HTML}{FCEFE2}

\newcommand{\fs}[1]{\fontsize{#1}{\dimexpr #1pt*6/5\relax}\selectfont}

\begin{figure*}[ht]
  \centering
 
\resizebox{\textwidth}{!}{%
\begin{tikzpicture}[
    x=1pt, y=-1pt,   %
    arr/.style={draw=arrowc, line width=1.1pt,
                -{Triangle[length=4.4pt,width=4.6pt]}},
    conn/.style={draw=arrowc, line width=1.1pt},
  ]
 
\newcommand{\search}[7]{%
  \draw[fill=#4bg, draw=#4, line width=1.2pt, rounded corners=3.6pt]
    (#1,#2) rectangle ++(#3,17.3);
  \pgfmathsetmacro{\magr}{#7/2}
  \pgfmathsetmacro{\magh}{#7*0.36}
  \draw[draw=#4, line width=1.4pt, line cap=round]
    (#1+3.9+#7*0.85,#2+3.4+#7*0.85) -- ++(\magh,\magh);
  \draw[fill=white, draw=#4, line width=1.1pt]
    (#1+3.9+\magr,#2+3.3+\magr) circle[radius=\magr];
  \node[anchor=west, inner sep=0pt, text=#4, font=\fs{#6}\bfseries]
    at (#1+3.9+#7+4.2,#2+8.65) {#5};}
\newcommand{\webpage}[5]{%
  \draw[fill=pagebg, draw=pageln, line width=0.9pt, rounded corners=2.16pt]
    (#1,#2) rectangle ++(#3,17.3);
  \fill[white] (#1+3.6,#2+3.6) rectangle ++(#4,2.2);
  \fill[white] (#1+3.6,#2+9.35) rectangle ++(#5,2.2);}
\newcommand{\agent}[6]{%
  \draw[fill=agentbg, draw=agentln, line width=1.3pt, rounded corners=3.6pt]
    (#1,#2) rectangle ++(#3,#4);
  \node[text=agentln, font=\fs{#5}\bfseries, inner sep=0pt]
    at (#1+#3/2,#2+#4/2) {#6};}
\newcommand{\prompt}[2]{%
  \draw[draw=frame, line width=1pt, rounded corners=3.6pt]
    (#1,#2) rectangle ++(432.8,34.6);
  \node[text=txt, font=\fs{11.5}\bfseries\itshape, align=center, inner sep=0pt]
    at (#1+216.4,#2+17.3)
    {Report the release year of: To Kill a Mockingbird, The Lord of the Rings,\\
     The Hobbit, Katabasis and The Book Thief.};}
\newcommand{\answer}[3]{%
  \draw[fill=answerbg, draw=frame, line width=1pt, rounded corners=3.6pt]
    (#1,#2) rectangle ++(432.8,34.6);
  \node[text=txt, font=\fs{#3}, inner sep=0pt] at (#1+216.4,#2+17.3)
    {1960, 1954, 1937, 2025, 2005};}
 
\node[font=\fs{18}\bfseries, inner sep=0pt] at (235.2,44.9) {Single Agent};
\prompt{10.1}{65.1}
 
\agent{40.8}{155.2}{114.8}{19.7}{13}{Single Agent}
 
\foreach \y/\lab/\pw/\la/\lb in {
    119.6/To Kill a Mockingbird/72.0/64.8/46.7,
    146.2/The Lord of the Rings/97.2/90.0/64.8,
    172.9/The Hobbit/57.6/50.4/36.3,
    199.5/Katabasis/111.6/104.4/75.2,
    226.2/The Book Thief/79.2/72.0/51.9} {
  \search{183.0}{\y}{104.4}{acc}{\lab}{7.8}{10.8}
  \draw[draw=arrowc, line width=1pt, -{Triangle[length=4pt,width=4.4pt]}]
    (287.4,\y+8.65) -- (293.9,\y+8.65);
  \webpage{293.9}{\y}{\pw}{\la}{\lb}
}
\foreach \y in {136.9,163.5,190.1,216.8} \draw[arr] (235.2,\y) -- ++(0,9.4);
\draw[arr] (235.2,243.4) -- (235.2,289.9);
 
\draw[arr] (235.4,107.6) -- (235.4,119.8); %
 
\answer{10.1}{303.8}{12}
 
\node[font=\fs{18}\bfseries, inner sep=0pt] at (673.4,48.5) {Delegation};
\prompt{460.2}{65.9}
 
\agent{567.9}{112.5}{211.0}{19.7}{13}{Orchestrator}
\draw[arr] (673.4,104.8) -- (673.4,112.0);
\draw[conn] (673.4,132.2) -- (673.4,140.9);
\draw[conn, line cap=round] (507.1,140.9) -- (839.8,140.9);
 
\foreach \i/\cx/\lab/\pw/\la/\lb/\yr in {
    1/507.1/Mockingbird/46.7/39.5/28.4/1960,
    2/590.2/LotR/58.0/50.8/36.6/1954,
    3/673.4/Hobbit/37.3/30.1/21.7/1937,
    4/756.6/Katabasis/60.0/52.8/38.9/2025,
    5/839.8/Book Thief/51.3/44.1/31.8/2005} {
  \draw[arr] (\cx,140.9) -- ++(0,7.9);
  \draw[fill=agentbg, draw=agentln, line width=1.3pt, rounded corners=3.6pt]
    (\cx-36,148.8) rectangle (\cx+36,245.0);
  \node[text=agentln, font=\fs{9}\bfseries, inner sep=0pt] at (\cx,157.8) {Executor \i};
  \search{\cx-33}{166.5}{66}{acc}{\lab}{7.5}{10.1}
  \draw[arr] (\cx,183.8) -- (\cx,191.0);
  \webpage{\cx-\pw/2}{191.0}{\pw}{\la}{\lb}
  \draw[arr] (\cx,208.3) -- (\cx,215.5);
  \draw[fill=white, draw=yearln, line width=1.1pt, rounded corners=2.9pt]
    (\cx-18,215.5) rectangle ++(36,21.6);
  \node[text=yeartx, font=\fs{9.5}\bfseries, inner sep=0pt] at (\cx,226.3) {\yr};
  \draw[arr] (\cx,245.0) -- (\cx,259.0);   %
}
 
\agent{471.0}{259.0}{404.8}{19.7}{13}{Orchestrator}
\draw[arr] (673.4,278.7) -- (673.4,289.9);
 
\answer{460.2}{303.8}{12}

\def\T{60}\def\C{9}
\newcommand{\axes}[3]{%
  \draw[arrowc, line width=0.8pt, -{Triangle[length=3pt,width=3pt]}] (#1,#2) -- ++(#3,0)
     node[anchor=north east, text=txt, font=\fs{9}, inner sep=1.5pt, pos=1] {time};
  \draw[arrowc, line width=0.8pt, -{Triangle[length=3pt,width=3pt]}] (#1,#2) -- ++(0,-58)
     node[anchor=south, text=txt, font=\fs{9}, inner sep=1.5pt] {context};}
 
\axes{40}{420}{390}
\foreach \i in {1,...,5}
  \draw[fill=notebg, draw=note, line width=0.8pt]
    (40+\i*\T-\T,420) rectangle ++(\T,-\i*\C);
\foreach \i in {1,...,5}
  \node[text=note, font=\fs{8}\bfseries, inner sep=0pt] at (40+\i*\T-\T/2,420-\i*\C/2) {\i};
\draw[note, dashed, line width=0.6pt] (40,420-5*\C) -- (40+4*\T,420-5*\C);
\node[anchor=east, text=note, font=\fs{9}\bfseries, inner sep=1pt] at (38,420-5*\C) {5};
\node[anchor=west, text=txt, font=\fs{9.5}, inner sep=0pt, align=left] at (40+5*\T+8,390)
  {\textbf{5 sequential steps}\\[1pt]peak context: 5 results};
 
\axes{490}{420}{390}
\def\D{14}   %
\draw[fill=agentbg, draw=agentln, line width=0.8pt] (490,420) rectangle ++(\D,-4.5);
\draw[fill=notebg, draw=note, line width=0.8pt] (490+\D,420) rectangle ++(\T,-\C);
\node[text=note, font=\fs{8}\bfseries, inner sep=0pt] at (490+\D+\T/2,420-\C/2) {E1--E5};
\node[anchor=south, text=txt, font=\fs{9}, inner sep=0pt] at (490+\D+\T/2,420-\C-3) {$\times$5 in parallel};
\node[anchor=east, text=note, font=\fs{9}\bfseries, inner sep=1pt] at (488,420-\C) {1};
\draw[fill=agentbg, draw=agentln, line width=0.8pt] (490+\D+\T,420) rectangle ++(\D,-4.5);
\node[anchor=north, text=agentln, font=\fs{8.5}, inner sep=0pt] at (490+\D/2,423) {dispatch};
\node[anchor=north, text=agentln, font=\fs{8.5}, inner sep=0pt] at (490+1.5*\D+\T,423) {aggregate};
\node[anchor=east, text=txt, font=\fs{9.5}, inner sep=0pt, align=left] at (490+390,390)
  {\textbf{1 parallel step}\\[1pt]peak context: 1 result per agent};
 
\draw[fill=notebg, draw=note, line width=0.8pt] (215,448) rectangle ++(12,-7);
\node[anchor=west, text=txt, font=\fs{10}, inner sep=0pt] at (231,444.6)
  {results in an agent's context window during a search step};
\draw[fill=agentbg, draw=agentln, line width=0.8pt] (480,446.5) rectangle ++(12,-4.5);
\node[anchor=west, text=txt, font=\fs{10}, inner sep=0pt] at (496,444.6)
  {orchestrator dispatching queries / aggregating responses};
\end{tikzpicture}%
}

\caption{
    \textit{Using a single agent vs. delegation for a composite task. }
    A single agent answers questions sequentially (\textit{top left}), calling the \texttt{\small web\_search} tool for each book and %
    accumulating outputs in a single context.
    Because these outputs may span several web pages, answers about a book at position $i$ attend over information retrieved for earlier books (\textit{bottom left}).
    Delegation instead partitions questions across executors (\textit{top right}), reducing peak context and unnecessary attention interactions (\textit{bottom right}).
    }
\label{fig:fig1_setup}
\end{figure*}
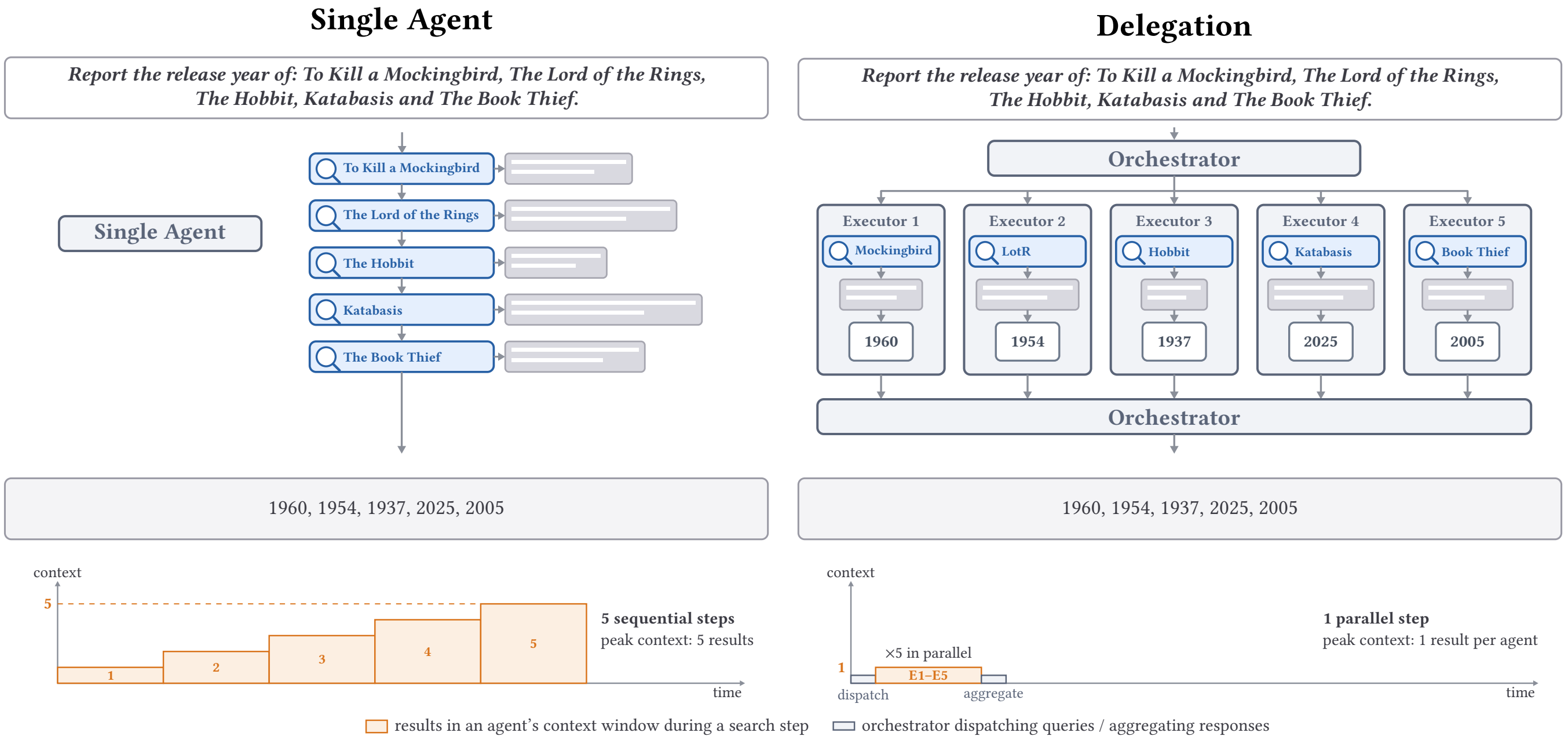

To answer this question, we construct four purpose-built  datasets that require retrieving facts via web search. These datasets require a search system to retrieve
(i) release years of books,
(ii) ages of Nobel Prize winners,
(iii) GDP growth rates, and
(iv) launch years of commercial products.
Together, these datasets cover popular and relatively niche facts.

Our dataset construction also accounts for a crucial distinction between LLM-based and traditional search: while traditional search always retrieves information from external sources, LLMs can answer many  questions from their parametric knowledge alone~\cite{kirsten2026characterizing,DBLP:conf/emnlp/PetroniRRLBWM19}. This parametric knowledge can influence whether delegation is needed at all, and how retrieved information is used. With this insight, we construct two versions of each dataset: one concerning facts before the underlying LLM's knowledge cutoff (\eg, books released before the cutoff), and  one about post-cutoff facts.
We compare  delegation against single-agent baselines using open- and closed source models.

Our experiments reveal that, surprisingly, \textbf{delegation does \textit{not} lead to a consistent increase in accuracy over a single-agent setup}. In fact, more often than not, delegation reduces accuracy. On the post-cutoff datasets, these accuracy drops for delegation are more pronounced. In contrast, single-agent accuracy is comparatively stable on average across the pre- and post-cutoff splits. This difference indicates a fundamental disparity in how single-agent and delegated setups process information  unknown to the underlying LLMs.

Deeper analysis sheds light on causes of these failures. 
First, orchestrator LLMs often inject additional context into delegated queries.
In a vast fraction of post-cutoff cases, this \textbf{added information is incorrect, steering the executor toward the wrong task}.
For instance,  when searching for the release year of the book \textit{The Calamity Club}, the orchestrator assumes it is searching for a \textit{film} and phrases the query as \textit{``Provide the release year of the most notable film or short film titled exactly 'The Calamity Club'...``}.

Second, \textbf{delegation introduces an information boundary, creating a disconnect between the context of the orchestrator and the executor}. Across this \uline{\textbf{delegation boundary}}, critical information, like retrieval failures, can be lost unless explicitly communicated.  For instance, when looking up GDP growth, the orchestrator LLM instructs the executor to retrieve information from the World Bank website. The  executor LLM fails to open a webpage due to a no-robots policy. This failure is not reported to the orchestrator, leading to repeated requests that encounter the same obstacle.

The delegation boundary also reduces the amount of trust the orchestrator places in the executor's research, despite instructions stating that the executor specializes in answering factual questions. For instance, an orchestrator trace shows the reasoning: ``The subagent says 2016, but I think `A Gathering of
Shadows' was published in 2022 [...]. Subagent is wrong.''

Third, \textbf{models sometimes use tools suboptimally}, packing multiple subtasks into a single executor query, making the executor's task more difficult. Other times, models issue queries that are simple rephrasings of previous queries, adding little new information. Both behaviors are more prevalent in delegation setups than in single-agent setups.

Mitigation experiments instructing executors to report retrieval failures to the orchestrator, or prohibiting models from issuing semantically equivalent queries lead to limited gains.

Our findings uncover both inherent tradeoffs and potentially remediable issues in delegation.
Using subagents limits context growth, but requires explicit communication of useful evidence and operational feedback across the delegation boundary, which the orchestrator must then appropriately incorporate. Other issues, such as models introducing unsupported information into delegated queries, are not unique to delegation, and may be amenable to targeted training. Such interventions, however, would need to generalize across domains and orchestrator--executor pairings.

\section{Related Work}

\xhdr{Multi-agent Orchestration.}
Multi-agent systems distribute tasks across interacting LLM agents, often using a central orchestrator to decompose requests, assign subtasks, and aggregate responses~\citep{hong2023metagpt,qian2023chatdev, shen2023hugginggpt}. 
We point to~\citep{guo2024large} for a broader survey of the design space.
Delegating tasks to subagents can have several motivations:
Decomposition of complex tasks~\citep{shen2023hugginggpt,hong2023metagpt, zhang2026orchestra},
specialization~\citep{li2023camel,fourney2024magentic}, parallel execution~\citep{kim2023llm,amayuelas2025self}, and
context distribution across agents~\citep{zhang2024chain, anthropic2025multiagent}.
However, prior studies find that multi-agent setups often fail to outperform single-agent baselines~\citep{zhang2025stop,tran2026single}. %
\citet{kim2026capable} find that the benefits of coordination depend on task structure and model capability, with overhead potentially outweighing possible gains.
\citet{zywot2026can} show that orchestrated smaller agents can beat a larger single agent on tool-intensive tasks, and that reasoning in subagents has only limited effects. %
\citet{cemri2025multi} identify common multi-agent failure modes, involving specification, agent coordination, and verification.
Complementing these studies, we isolate the effect of delegating tool access on performance.

\xhdr{Knowledge-aware Tool Use.}
Retrieval augmentation and web search have emerged as popular tools to provide LLMs with access to external knowledge and update their static, potentially stale, parametric knowledge~\citep{lewis2020retrieval,vu2024freshllms}.
Recent work proposed different adaptive retrieval strategies to decide when and what to retrieve from the web, using confidence estimates~\citep{jiang2023active,wang2023self}, question difficulty~\citep{jeong2024adaptive}, or time sensitivity~\citep{vu2024freshllms}.
Yet, models do not reliably know when they need to retrieve~\citep{ren2025investigating,liu2026busy}, and redundant retrieval can make models less efficient or even degrade accuracy~\citep{wu2026call,maekawa2024retrieval}.
\citet{fan2026livebrowsecomp} show that search agents often rely on parametric knowledge even when using search tools, and perform worse on questions requiring recent facts.
While prior work largely studies retrieval decisions within a single agent, we examine how these decisions change under delegation.
By comparing pre- and post-knowledge-cutoff facts, we distinguish settings where parametric knowledge may suffice from those requiring external information.

\xhdr{Knowledge Conflicts and Information Transfer.}
Access to external information does not guarantee that models use it correctly.
Models may favor memorized answers over retrieved evidence~\citep{longpre2021entity}, or base their decision on how evidence is presented~\citep{xie2023adaptive}.
Temporal knowledge conflicts also arise when models prefer outdated facts despite encoding more recent information~\citep{hossain2026right}.
Prior work primarily examines how individual models reconcile parametric knowledge with external evidence.
In multi-agent systems, information must additionally pass between agents, introducing further opportunities for errors.
We investigate these interactions at the level of individual facts, tracing where information is lost or altered.

\xhdr{Long-Context and Long-Horizon Tasks.}
Many tasks require processing more information or executing more steps than what fits in a single model's context.
Such demands may arise from depth (long reasoning or action sequences), or breadth (retrieval and synthesis across many sources).
Prior work addresses these challenges through context compression, persistent state, and recursive model calls~\citep{wu2025resum,pmlr-v306-kang26,zhang2026recursivelanguagemodels}.
Multi-agent systems instead distribute information gathering and processing across agents, reducing individual context requirements and enabling parallelization~\citep{anthropic2025multiagent,zhang2024chain,zhao-etal-2024-longagent}.
Our work complements these efforts by studying the reliability of delegation in acquiring and integrating external information.

\section{Experimental Setup} %
\label{sec:exp-setup}
We study the effectiveness of delegation across four agentic configurations, four models, and four factual datasets.
Each dataset includes pre- and post-cutoff conditions to examine how delegation behaves when the requested information is more or less likely to be available in the model's parametric knowledge.

\xhdr{Model-tool configurations.}
We consider four configurations, illustrated in~\Cref{fig:agent-setups} in the Appendix:

\noindent
\begin{enumerate*}[label=(\arabic*), itemjoin={{; }}, itemjoin*={{; and }}]
    \item \textbf{No tool:} A single LLM  answers the query without access to external tools, serving as a closed-book baseline
    \item \textbf{Single agent:} A single LLM answers the query with direct access to web search
    \item \textbf{Forced delegation:} An orchestrator decomposes the query into atomic subtasks and delegates each to an executor (an LLM with access to web search). The orchestrator then combines the executor responses into a final answer
    \item \textbf{Optional delegation:} The orchestrator may answer each subtask directly or delegate it to a tool-enabled executor.
\end{enumerate*}
In both delegation setups, the orchestrator has no direct access to web search.
Orchestrators and executors use the same underlying model but operate in separate contexts.
Each executor starts with a fresh context, receives a subtask formulated by the orchestrator, processes it, and returns its final answer.
The orchestrator only receives these final answers, not the executors' full trajectories and tool-use traces. This setup prevents context bloat (see \Cref{fig:fig1_setup}).

\noindent
\textit{Rationale behind model-tool configurations.}
These configurations are designed to separate different aspects of delegation, allowing us to compare the benefit of external information (no tool vs. single agent), the effect of mediating retrieval through another agent (single agent vs. forced delegation), and the ability to selectively delegate subtasks (forced vs. optional delegation).
The system instructions for each configuration are provided in Table~\ref{tab:agent_prompts}.

\xhdr{Models and Tools.} We evaluate four models: Claude Sonnet 4.5 (\claude), GPT-5 (\gpt), gpt-oss-120b (\gptoss), and Qwen3-30B-A3B-Thinking-2507 (\qwen). 
These represent different model families, including both closed- and open-source models that support tool use and instruction following.
Reported knowledge cutoffs range from June 2024 to July 2025. Model identifiers, providers, and cutoff dates are listed in Table~\ref{tab:models} in Appendix~\ref{app:model_details}.
All tool-enabled agents use the same web search interface, which retrieves Google search results through SerpAPI.\footnote{\url{https://serpapi.com/search-api}} Our interface exposes two operations: \texttt{search}, which returns a list of result titles, URLs, and snippets, and \texttt{open\_url}, which returns extracted page content for selected results as truncated Markdown.

\xhdr{Datasets.}
We construct four factual datasets in which each query requests multiple independently verifiable facts (see Figure~\ref{fig:fig1_setup} for an example).
This structure allows the same task to be solved either by a single agent or through delegation of individual subtasks.
All datasets involve time-dependent factual information.
Table~\ref{tab:dataset_overview} in Appendix~\ref{app:datasets} summarizes dataset statistics and sources.

\noindent
\textit{\textbf{(i) Books.}} Each query asks for the original publication years of five books published in the same year. We sample titles from annual Goodreads Top 100 lists for 2014--2019 and 2026, constructing 20 queries per year (120 pre-cutoff, 20 post-cutoff).
After manually reviewing all titles to verify publication years, we excluded 15 books whose original publication years differ from their list years, leaving 585 pre-cutoff and 100 post-cutoff books for evaluation.

\noindent
\textit{\textbf{(ii) Nobel Prize.}} Each query asks for the ages of all individual laureates awarded a Nobel Prize when they received the award in a specified category and year.
We construct queries across all six prize categories for the years 2014--2019 and 2025--2026, using the Nobel Prize API.
We exclude prizes awarded to organizations and include all individual laureates. %

\noindent
\textit{\textbf{(iii) GDP.}} Each query asks for the annual real GDP growth rates of four countries in a specified year, rounded to one decimal place. We obtain values from the World Bank API (indicator \texttt{\Small NY.GDP.MKTP.KD.ZG}) for 2019 (pre-cutoff) and 2025 (post-cutoff).
We randomly select five countries from each of five continents (Africa, Asia, Europe, North America and South America) with data available for both years.
From the 25 countries, we construct 30 distinct four-country sets through repeated shuffling, with each country appearing in four or five prompts. We use the same country sets in pre- and post-cutoff conditions, so that prompts differ only in the year.

\noindent
\textit{\textbf{(iv) Products.}} Each query asks for the launch years of four products recognized by the iF Design Award in the Product Design discipline. We collect winners from the iF Design Award website and retain products launched in 2021 (awarded in 2022) or 2026 (awarded in 2026). For each period, we select 20 products and construct 30 prompts by randomly sampling distinct sets of four products.

\xhdr{Pre- and Post-Cutoff Conditions.}
To examine whether delegation behaves differently depending on the likely availability of factual knowledge, we divide each dataset into pre-cutoff and post-cutoff splits.
Each dataset contains a variable that allows for such splitting. These variables are:
\textit{publication year} for Books, \textit{award year} for Nobel Prizes, \textit{growth rate in a year} for GDP and \textit{launch year} for Products. The splits are:
\begin{enumerate*}[label=(\arabic*), itemjoin={{; }}, itemjoin*={{; and }}]
\item \textbf{Pre-Cutoff:} Facts from 2014--2019 (Books, Nobel Prize, GDP) or 2021 (Products), preceding the reported knowledge cutoffs of all evaluated models
\item \textbf{Post-Cutoff:} Facts from the most recent available year: 2025 (GDP), 2026 (Books, Products), or 2025--2026 (Nobel Prize), which became available after the reported knowledge cutoffs.
\end{enumerate*}

\xhdr{Metrics.}
We measure accuracy at the prompt level. Each prompt asks about several items (books, countries, laureates or products). For instance, a prompt from the Books dataset asks: \textit{Determine the release year for each of the following:...}, followed by a list of book titles.
The accuracy on a dataset is computed as follows:
\[
\mathrm{Acc} = \frac{1}{N} \sum_{i=1}^{N} \frac{1}{M_i}\Bigg(\sum_{j=1}^{M_i} \mathbf{1}_{ij}\Bigg),
\]
where \(N\) is the number of prompts in the dataset, \(M_i\) is the number of sub-questions (items) in each prompt, and 
the indicator variable \(\mathbf{1}_{ij}\) indicates whether item \(j\) of prompt \(i\) is answered correctly. For GDP, an answer counts as correct if it is within 0.1 percentage points of the ground truth.
We report 95\% percentile bootstrap confidence intervals, resampling prompts with replacement (\(B = 1000\)), recomputing \(\mathrm{Acc}\) on each resample.

\xhdr{Tool Call Limits.} 
Each model operates under a fixed tool call budget that scales with the number of items $n$ in each query.
In the single-agent setup, the agent may make up to $2n$ \texttt{\small web\_search} calls and $3n$ \texttt{\small open\_url} calls. In both delegation setups, the orchestrator may make up to $2n$ executor calls, corresponding to the $2n$ \texttt{web\_search} calls for the single agent. Each executor receives a fresh budget of two \texttt{web\_search} calls and three \texttt{open\_url} calls.
Budgets are specified in the agents' system prompts and enforced during execution. Failed tool calls because of errors outside the model's control, such as failed search requests, do not count toward the limit.
Exact budgets and further implementation details are provided in Table~\ref{tab:tool_call_limits} in Appendix~\ref{app:tool-call-limits}.

\section{Delegation Often Reduces Accuracy}

\begin{table}[t]
    \centering
        \begin{tabular}{lccc}
            \toprule
            \textbf{Setup} &
            \textbf{Pre-cutoff} &
            \textbf{Post-cutoff} &
            \textbf{Change} \\
            \midrule

            No Tools
            & 0.53 & 0.01 & \dab{98\%} \\

            Single Agent
            & 0.77 & 0.75 & \dab{3\%} \\

            Optional Del.
            & 0.74 & 0.64 & \dab{14\%} \\

            Forced Del.
            & 0.77 & 0.66 & \dab{13\%} \\

            \bottomrule
        \end{tabular}%
    \caption{\textit{Accuracy drops from pre- to post-cutoff splits are larger for delegated setups than for the single agent.}
    Average accuracy across pre- and post-cutoff splits.
    }
    \label{tab:cutoff-accuracy-summary}
\end{table}

\begin{table*}[t]
    \centering
    \small
    \setlength{\tabcolsep}{3.5pt}

    \begin{subtable}{\linewidth}
        \centering
        \resizebox{\linewidth}{!}{%
            \begin{tabular}{lllllllll}
                \toprule
                \textbf{Books}
                & \multicolumn{4}{c}{\textbf{Pre-cutoff} ($n=585, N = 120$)}
                & \multicolumn{4}{c}{\textbf{Post-cutoff} ($n=100, N = 20$)} \\
                \cmidrule(r){2-5}
                \cmidrule(l){6-9}
                \textbf{Model} &
                \textbf{Single Agent} &
                \textbf{Optional Del.} &
                \textbf{Forced Del.} &
                \textbf{No Tools} &
                \textbf{Single Agent} &
                \textbf{Optional Del.} &
                \textbf{Forced Del.} &
                \textbf{No Tools} \\
                \midrule

                \claude
                & 0.85 {\scriptsize [0.80, 0.89]}
                & 0.83 {\scriptsize [0.79, 0.87]} \dab{2\%}
                & 0.83 {\scriptsize [0.79, 0.87]} \dab{2\%}
                & 0.75 {\scriptsize [0.69, 0.80]} \dab{12\%}
                & 0.90 {\scriptsize [0.84, 0.95]}
                & 0.80 {\scriptsize [0.70, 0.89]} \dab{11\%}
                & 0.81 {\scriptsize [0.69, 0.91]} \dab{10\%}
                & 0.00 {\scriptsize [0.00, 0.00]} \dab{100\%} \\

                \gpt
                & 0.95 {\scriptsize [0.92, 0.97]}
                & 0.92 {\scriptsize [0.89, 0.96]} \dab{3\%}
                & 0.93 {\scriptsize [0.89, 0.96]} \dab{2\%}
                & 0.93 {\scriptsize [0.89, 0.96]} \dab{3\%}
                & 0.91 {\scriptsize [0.80, 1.00]}
                & 0.52 {\scriptsize [0.37, 0.66]} \dab{43\%}
                & 0.60 {\scriptsize [0.46, 0.72]} \dab{34\%}
                & 0.00 {\scriptsize [0.00, 0.00]} \dab{100\%} \\

                \gptoss
                & 0.74 {\scriptsize [0.68, 0.79]}
                & 0.60 {\scriptsize [0.54, 0.66]} \dab{18\%}
                & 0.63 {\scriptsize [0.57, 0.69]} \dab{14\%}
                & 0.26 {\scriptsize [0.21, 0.30]} \dab{65\%}
                & 0.85 {\scriptsize [0.78, 0.91]}
                & 0.60 {\scriptsize [0.45, 0.73]} \dab{29\%}
                & 0.62 {\scriptsize [0.49, 0.75]} \dab{27\%}
                & 0.00 {\scriptsize [0.00, 0.00]} \dab{100\%} \\

                \qwen
                & 0.55 {\scriptsize [0.50, 0.60]}
                & 0.52 {\scriptsize [0.46, 0.58]} \dab{5\%}
                & 0.58 {\scriptsize [0.52, 0.64]} \uag{6\%}
                & 0.36 {\scriptsize [0.32, 0.41]} \dab{34\%}
                & 0.70 {\scriptsize [0.56, 0.83]}
                & 0.56 {\scriptsize [0.39, 0.72]} \dab{20\%}
                & 0.65 {\scriptsize [0.51, 0.77]} \dab{7\%}
                & 0.00 {\scriptsize [0.00, 0.00]} \dab{100\%} \\

                \bottomrule
            \end{tabular}%
        }
    \end{subtable}

    \vspace{0.35em}

    \begin{subtable}{\linewidth}
        \centering
        \resizebox{\linewidth}{!}{%
            \begin{tabular}{lllllllll}
                \toprule
                \textbf{GDP}
                & \multicolumn{4}{c}{\textbf{Pre-cutoff} ($n=120, N = 30$)}
                & \multicolumn{4}{c}{\textbf{Post-cutoff} ($n=120,  N= 30$)} \\
                \cmidrule(r){2-5}
                \cmidrule(l){6-9}
                \textbf{Model} &
                \textbf{Single Agent} &
                \textbf{Optional Del.} &
                \textbf{Forced Del.} &
                \textbf{No Tools} &
                \textbf{Single Agent} &
                \textbf{Optional Del.} &
                \textbf{Forced Del.} &
                \textbf{No Tools} \\
                \midrule

                \claude
                & 0.68 {\scriptsize [0.58, 0.77]}
                & 0.70 {\scriptsize [0.62, 0.78]} \uag{4\%}
                & 0.69 {\scriptsize [0.60, 0.77]} \uag{3\%}
                & 0.43 {\scriptsize [0.34, 0.53]} \dab{36\%}
                & 0.68 {\scriptsize [0.60, 0.75]}
                & 0.73 {\scriptsize [0.68, 0.80]} \uag{9\%}
                & 0.73 {\scriptsize [0.68, 0.80]} \uag{9\%}
                & 0.01 {\scriptsize [0.00, 0.03]} \dab{99\%} \\

                \gpt
                & 1.00 {\scriptsize [1.00, 1.00]}
                & 0.97 {\scriptsize [0.92, 1.00]} \dab{3\%}
                & 0.98 {\scriptsize [0.94, 1.00]} \dab{3\%}
                & 0.38 {\scriptsize [0.28, 0.47]} \dab{63\%}
                & 0.89 {\scriptsize [0.79, 0.98]}
                & 0.43 {\scriptsize [0.33, 0.53]} \dab{52\%}
                & 0.35 {\scriptsize [0.25, 0.45]} \dab{61\%}
                & 0.06 {\scriptsize [0.03, 0.10]} \dab{93\%} \\

                \gptoss
                & 0.92 {\scriptsize [0.83, 0.98]}
                & 0.65 {\scriptsize [0.55, 0.73]} \dab{29\%}
                & 0.73 {\scriptsize [0.63, 0.82]} \dab{21\%}
                & 0.18 {\scriptsize [0.11, 0.26]} \dab{81\%}
                & 0.77 {\scriptsize [0.66, 0.87]}
                & 0.70 {\scriptsize [0.61, 0.78]} \dab{9\%}
                & 0.73 {\scriptsize [0.65, 0.80]} \dab{5\%}
                & 0.06 {\scriptsize [0.02, 0.11]} \dab{92\%} \\

                \qwen
                & 0.30 {\scriptsize [0.22, 0.38]}
                & 0.28 {\scriptsize [0.20, 0.35]} \dab{8\%}
                & 0.34 {\scriptsize [0.26, 0.43]} \uag{14\%}
                & 0.23 {\scriptsize [0.14, 0.31]} \dab{25\%}
                & 0.39 {\scriptsize [0.31, 0.48]}
                & 0.39 {\scriptsize [0.30, 0.48]}
                & 0.44 {\scriptsize [0.35, 0.54]} \uag{13\%}
                & 0.05 {\scriptsize [0.01, 0.10]} \dab{87\%} \\

                \bottomrule
            \end{tabular}%
        }
    \end{subtable}

    \vspace{0.35em}

    \begin{subtable}{\linewidth}
        \centering
        \resizebox{\linewidth}{!}{%
            \begin{tabular}{lllllllll}
                \toprule
                \textbf{Nobel}
                & \multicolumn{4}{c}{\textbf{Pre-cutoff} ($n=72, N = 34$)}
                & \multicolumn{4}{c}{\textbf{Post-cutoff} ($n=24, N = 12$)} \\
                \cmidrule(r){2-5}
                \cmidrule(l){6-9}
                \textbf{Model} &
                \textbf{Single Agent} &
                \textbf{Optional Del.} &
                \textbf{Forced Del.} &
                \textbf{No Tools} &
                \textbf{Single Agent} &
                \textbf{Optional Del.} &
                \textbf{Forced Del.} &
                \textbf{No Tools} \\
                \midrule

                \claude
                & 0.88 {\scriptsize [0.78, 0.95]}
                & 0.87 {\scriptsize [0.77, 0.96]} \dab{1\%}
                & 0.94 {\scriptsize [0.88, 0.98]} \uag{7\%}
                & 0.89 {\scriptsize [0.79, 0.97]} \uag{2\%}
                & 1.00 {\scriptsize [1.00, 1.00]}
                & 0.90 {\scriptsize [0.76, 1.00]} \dab{10\%}
                & 0.88 {\scriptsize [0.74, 1.00]} \dab{13\%}
                & 0.00 {\scriptsize [0.00, 0.00]} \dab{100\%} \\

                \gpt
                & 0.94 {\scriptsize [0.87, 0.99]}
                & 0.94 {\scriptsize [0.87, 0.99]}
                & 0.94 {\scriptsize [0.87, 0.99]}
                & 0.93 {\scriptsize [0.86, 0.98]} \dab{1\%}
                & 0.93 {\scriptsize [0.82, 1.00]}
                & 0.93 {\scriptsize [0.82, 1.00]}
                & 0.90 {\scriptsize [0.79, 1.00]} \dab{3\%}
                & 0.00 {\scriptsize [0.00, 0.00]} \dab{100\%} \\

                \gptoss
                & 0.94 {\scriptsize [0.85, 0.99]}
                & 0.89 {\scriptsize [0.80, 0.97]} \dab{5\%}
                & 0.91 {\scriptsize [0.83, 0.97]} \dab{4\%}
                & 0.87 {\scriptsize [0.78, 0.95]} \dab{7\%}
                & 0.88 {\scriptsize [0.74, 1.00]}
                & 0.85 {\scriptsize [0.67, 1.00]} \dab{3\%}
                & 0.82 {\scriptsize [0.60, 1.00]} \dab{6\%}
                & 0.00 {\scriptsize [0.00, 0.00]} \dab{100\%} \\

                \qwen
                & 0.82 {\scriptsize [0.71, 0.92]}
                & 0.92 {\scriptsize [0.83, 0.98]} \uag{12\%}
                & 1.00 {\scriptsize [1.00, 1.00]} \uag{21\%}
                & 0.84 {\scriptsize [0.74, 0.94]} \uag{2\%}
                & 0.28 {\scriptsize [0.03, 0.53]}
                & 0.36 {\scriptsize [0.14, 0.64]} \uag{30\%}
                & 0.39 {\scriptsize [0.14, 0.64]} \uag{40\%}
                & 0.00 {\scriptsize [0.00, 0.00]} \dab{100\%} \\

                \bottomrule
            \end{tabular}%
        }
    \end{subtable}

    \vspace{0.35em}

    \begin{subtable}{\linewidth}
        \centering
        \resizebox{\linewidth}{!}{%
            \begin{tabular}{lllllllll}
                \toprule
                \textbf{Products}
                & \multicolumn{4}{c}{\textbf{Pre-cutoff} ($n=120, N = 30$)}
                & \multicolumn{4}{c}{\textbf{Post-cutoff} ($n=120, N = 30$)} \\
                \cmidrule(r){2-5}
                \cmidrule(l){6-9}
                \textbf{Model} &
                \textbf{Single Agent} &
                \textbf{Optional Del.} &
                \textbf{Forced Del.} &
                \textbf{No Tools} &
                \textbf{Single Agent} &
                \textbf{Optional Del.} &
                \textbf{Forced Del.} &
                \textbf{No Tools} \\
                \midrule

                \claude
                & 0.73 {\scriptsize [0.64, 0.83]}
                & 0.83 {\scriptsize [0.75, 0.91]} \uag{14\%}
                & 0.89 {\scriptsize [0.83, 0.95]} \uag{22\%}
                & 0.25 {\scriptsize [0.17, 0.34]} \dab{66\%}
                & 0.82 {\scriptsize [0.73, 0.89]}
                & 0.74 {\scriptsize [0.66, 0.83]} \dab{9\%}
                & 0.79 {\scriptsize [0.72, 0.86]} \dab{3\%}
                & 0.00 {\scriptsize [0.00, 0.00]} \dab{100\%} \\

                \gpt
                & 0.89 {\scriptsize [0.83, 0.94]}
                & 0.84 {\scriptsize [0.78, 0.90]} \dab{6\%}
                & 0.81 {\scriptsize [0.73, 0.88]} \dab{9\%}
                & 0.64 {\scriptsize [0.54, 0.73]} \dab{28\%}
                & 0.87 {\scriptsize [0.79, 0.93]}
                & 0.68 {\scriptsize [0.59, 0.78]} \dab{21\%}
                & 0.65 {\scriptsize [0.56, 0.74]} \dab{25\%}
                & 0.03 {\scriptsize [0.00, 0.08]} \dab{96\%} \\

                \gptoss
                & 0.76 {\scriptsize [0.65, 0.85]}
                & 0.64 {\scriptsize [0.57, 0.72]} \dab{15\%}
                & 0.67 {\scriptsize [0.56, 0.77]} \dab{12\%}
                & 0.25 {\scriptsize [0.18, 0.33]} \dab{67\%}
                & 0.78 {\scriptsize [0.65, 0.88]}
                & 0.63 {\scriptsize [0.56, 0.69]} \dab{19\%}
                & 0.68 {\scriptsize [0.61, 0.75]} \dab{13\%}
                & 0.00 {\scriptsize [0.00, 0.00]} \dab{100\%} \\

                \qwen
                & 0.38 {\scriptsize [0.28, 0.48]}
                & 0.40 {\scriptsize [0.33, 0.49]} \uag{7\%}
                & 0.37 {\scriptsize [0.26, 0.48]} \dab{2\%}
                & 0.30 {\scriptsize [0.21, 0.40]} \dab{20\%}
                & 0.33 {\scriptsize [0.19, 0.47]}
                & 0.36 {\scriptsize [0.24, 0.48]} \uag{10\%}
                & 0.58 {\scriptsize [0.47, 0.69]} \uag{77\%}
                & 0.00 {\scriptsize [0.00, 0.00]} \dab{100\%} \\

                \bottomrule
            \end{tabular}%
        }
    \end{subtable}

    \caption{\textit{Delegation performs worse than the single agent in most cases.}
    Accuracy across datasets, models, and delegation setups.
    Each entry reports accuracy with a 95\% prompt-level bootstrap confidence interval (CI). $N$ denotes the number of prompts and $n$ the number of scored items per dataset and cutoff period. 
    Arrows indicate relative accuracy changes from the corresponding
    single-agent baseline, computed from unrounded accuracies.}
    \label{tab:accuracy}
\end{table*}

We first compare accuracy across the four agent setups, models, and datasets (Table~\ref{tab:accuracy}).
We find that tool access is essential beyond the knowledge cutoff.
On post-cutoff tasks, models without tools achieve only $0$--$6\%$ accuracy, whereas single agents with direct tool access frequently reach $70$--$100\%$. Access to external information is therefore crucial for answering questions beyond the models' knowledge cutoff.
We continue to compare the performance of the delegated setups to the single agent as a baseline.

\xhdr{Delegation does not improve accuracy over direct tool use.}
Across all 32 model--dataset--cutoff combinations, optional delegation performs worse than the single agent in 22 cases (better in 7, tied in 3), while forced delegation performs worse in 21 (better in 10, tied in 1).
Thus, introducing additional agents generally provides no accuracy benefit.

\xhdr{Performance drops of delegated setups are larger on post-cutoff tasks.}
Averaged equally across models and datasets, single agent accuracy remains relatively stable between pre- and post-cutoff tasks ($0.77$ \vs $0.75$).
In contrast, optional and forced delegation both lose approximately $10$ percentage points (pp) in absolute accuracy (Table~\ref{tab:cutoff-accuracy-summary}).
Consequently, the gap to direct tool use increases from $1$ to $9$pp for forced delegation and from $3$ to $11$pp for optional delegation.
The largest losses occur for \gpt: on post-cutoff GDP prompts, accuracy drops significantly from $0.89$ to $0.43$ and $0.35$ under optional and forced delegation, respectively.
Similar differences arise on post-cutoff Books and Products.
Across datasets, Books exhibits the largest average post-cutoff losses in accuracy for delegation, while Nobel questions remain comparatively robust.

\xhdr{Forced delegation generally outperforms optional delegation.} 
Forced delegation achieves higher accuracy than optional delegation in most configurations, particularly for \gptoss and \qwen.
However, neither consistently outperforms the single agent.
Under optional delegation, the orchestrator decided to answer from memory for $21\%$ of items.
Allowing orchestrators to decide whether to delegate therefore does not reliably prevent performance losses. 
For example, on a pre-cutoff Books query, \gptoss incorrectly recalled the release year of the book ``The Room on Rue Amélie'' as \textit{2021} under optional delegation and attributed it to the wrong author, but retrieved the correct year (\textit{2018}) with forced delegation.

\xhdr{Delegation effects vary across models and setups.}
\gpt achieves the strongest single-agent performance ($0.87$--$1.00$), but suffers the largest post-cutoff accuracy drops for delegated setups, particularly on GDP (up to $54$pp). 
\claude slightly benefits from delegation on some pre-cutoff tasks ($+2$--$5$pp on average), but loses $5$--$6$pp post-cutoff.
\gptoss loses substantial accuracy under delegation even pre-cutoff, whereas \qwen benefits from forced delegation in 6 out of 8 conditions, gaining up to $25$pp on post-cutoff Products.

Overall, delegation frequently reduces accuracy despite providing access to the same external tools, particularly when answering questions beyond the knowledge cutoff.
We next examine where and how these failures emerge.

\section{Where Delegation Breaks Down}
\label{sec:delegation-breaks}

We identify recurring behavioral failure modes and examine whether they arise specifically from delegation or reflect broader limitations of agentic tool use.

\begin{figure}[th]
    \centering
    \includegraphics[width=\linewidth]{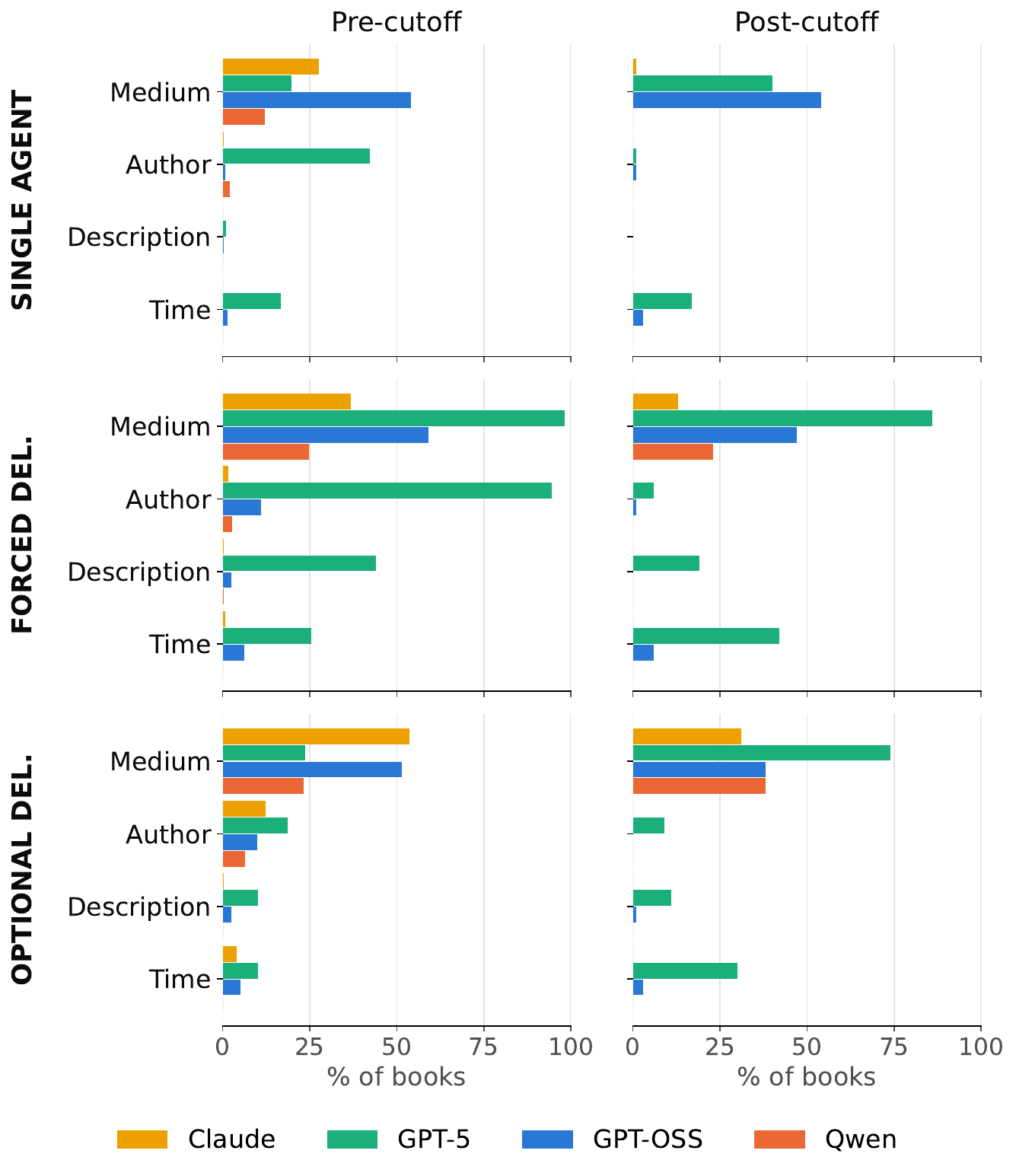}
    \caption{\textit{Models often add details about the item absent from the user prompt.}
    Bars show the share of books whose first executor prompt (single-agent: first \texttt{\small web\_search query}) adds information absent from the original user query.}
    \label{fig:addedbelief}
\end{figure}

\subsection{Orchestrators Inject Incorrect Assumptions} %
Orchestrators may alter requested information when formulating executor queries.
To analyze these changes, we use an LLM judge to identify attributes added or changed relative to the original request.

\xhdr{Orchestrators frequently enrich delegated queries.}
Figure~\ref{fig:addedbelief} reports the frequency and composition of query modifications on Books.
Common additions are the publication medium (e.g. \textit{novel} or \textit{book}) and author (e.g. \textit{Abby Jimenez}). 
Their frequency varies substantially across models and delegation settings.
In particular, \gpt adds the most information, specifying the medium in up to 98\% of pre-cutoff queries under forced delegation. Such additions may provide useful context, helping to disambiguate the requested information, but they also introduce assumptions about the intended entity or how the executor should select it.
\textbf{Injecting assumptions is more common for delegated setups.}
Notably, delegators add more information to their delegated queries than single agents add to their search queries both pre- and post-cutoff.

\xhdr{Added information is less reliable for post-cutoff items.}
Table~\ref{tab:books-medium-author-correct} evaluates the correctness of added author and medium information.
Most pre-cutoff author additions are correct.
However, delegation adds only incorrect author information on the post-cutoff split.
Medium identification follows a similar pattern.
Under 
delegation, only $4$--$6\%$ of added medium information is correct for \gpt on post-cutoff books.
Similar errors occur for \gptoss and \qwen.
Notably, \gpt adds author information much less frequently post-cutoff, but continues to introduce other attributes.
Thus, orchestrators add less identifying information for unfamiliar books, while the information they do add is often less reliable.
This pattern suggests that orchestrators can use parametric knowledge to enrich queries when they recognize the requested item, but cannot provide the same contextual information when the item is unfamiliar.

\xhdr{Incorrect assumptions can distort the delegated task.}
When added information is incorrect, the executor may be directed toward a different entity than the one requested.
For example, when asked about the release year of the 2026 book \textit{The Night We Met} by Abby Jimenez, \gpt delegated the prompt \textit{``Determine the release year (first public release) of the song "The Night We Met" by Lord Huron.''} The executor correctly returned 2015, the release year of the song. 
Here, the executor answers the delegated query correctly, yet produces an incorrect answer to the user's original request.
Therefore, query rewriting introduces an additional opportunity for error at the orchestrator--executor boundary. 
Incorrect additions can also persist even when the executor encounters contradicting evidence. For \textit{The Calamity Club}, a 2026 novel by Kathryn Stockett, \gpt delegates: \textit{``Provide the release year of the most notable film or short film titled exactly 'The Calamity Club'...``}. The executor encountered evidence for the correct 2026 book three times in the search results.
Nevertheless, it answers 1913 by guessing, which the orchestrator adopts as its final answer. 
Such an assumption can therefore anchor the executor's search and final response.

These examples showcase a failure point at the orchestrator--executor boundary: Query rewriting can alter the task itself. Especially for post-cutoff items, assumptions drawn from familiar entities can redirect retrieval away from the requested information, and encountering the correct source does not necessarily resolve the resulting error. This failure affects the single agent less, since even for wrong additions to its web search queries, it has access to previously made assumptions and corresponding search results, avoiding loss of information.

\begin{table}[t]
    \centering
    \small
    \setlength{\tabcolsep}{4pt}
    \begin{tabular}{@{}llcc@{\hspace{12pt}}cc@{}}
        \toprule
        & & \multicolumn{2}{c}{\textbf{Pre-cutoff}} & \multicolumn{2}{c}{\textbf{Post-cutoff}} \\
        \cmidrule(lr){3-4} \cmidrule(l){5-6}
        \textbf{Model} & \textbf{Setting} & Medium & Author & Medium & Author \\
        \midrule
        \gptoss & Single agent & 58 & 80 & 28 \dab{52\%} & 100 \uag{25\%} \\
         & Forced & 58 & 72 & 21 \dab{63\%} & 0 \dab{100\%} \\
         & Optional & 51 & 71 & 16 \dab{69\%} & -- \\
        \midrule
        \qwen & Single agent & 82 & 92 & -- & -- \\
         & Forced & 69 & 69 & 22 \dab{69\%} & -- \\
         & Optional & 81 & 60 & 53 \dab{35\%} & -- \\
        \midrule
        \gpt & Single agent & 91 & 96 & 63 \dab{32\%} & 0 \dab{100\%} \\
         & Forced & 93 & 95 & 6 \dab{94\%} & 0 \dab{100\%} \\
         & Optional & 76 & 84 & 4 \dab{95\%} & 0 \dab{100\%} \\
        \midrule
        \claude & Single agent & 99 & 100 & 0 \dab{100\%} & -- \\
         & Forced & 64 & 100 & 0 \dab{100\%} & -- \\
         & Optional & 70 & 99 & 0 \dab{100\%} & -- \\
        \bottomrule
    \end{tabular}
    \caption{\textit{For post-cutoff books, orchestrators add medium and author details, often incorrectly.} Cells show the percentage of correct additions when the first executor prompt (single-agent: first \texttt{\small web\_search} query) adds the field. Arrows show relative change from pre- to post-cutoff.}
    \label{tab:books-medium-author-correct}
\end{table}

\subsection{Inefficient Tool Use}

Our tasks provide a simple setting for examining how models organize tool use.
Each request concerns independent items, making item-level decomposition an obvious strategy:
For example, an orchestrator can delegate each book to a separate executor.
Within each item-level search, an unsuccessful query should motivate a reformulation that seeks different information or sources.
We examine whether models follow these two patterns, first within search trajectories and then across executor calls. Appendix \ref{app:add-results} provides complementary statistics on the average number of \texttt{\small web\_search} and \texttt{\small open\_url} calls across models and settings. Tool usage varies substantially across models with the largest increases from pre- to post-cutoff questions observed for \gpt and \gptoss.

\begin{figure}
    \centering
    \includegraphics[width=.9\linewidth]{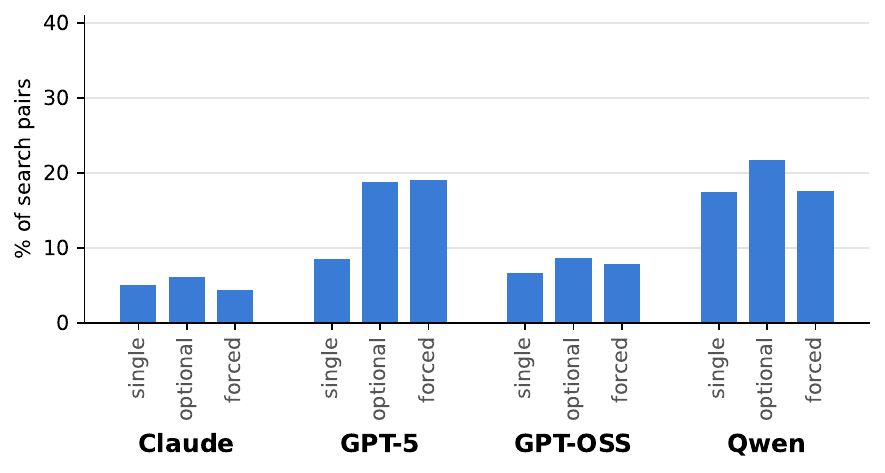}
    \caption{\textit{Models frequently repeat similar queries with only minor changes in wording.} Bars show the percentage of consecutive web\_search pairs classified as doing the same search, pooled across datasets.}
    \label{fig:querysimilarity}
\end{figure}

\xhdr{Models repeatedly issue redundant queries.}
Across tool-enabled settings, \textbf{40\%} of item-level search trajectories contain multiple web searches.
To assess whether repeated searches refine the original request, we compare consecutive queries concerning the same item.
We use LLM judges to identify newly introduced factual attributes and to decide whether two queries are the same search. We validate the similarity judge against three human annotators, finding substantial agreement
(Appendix \ref{sec:human_validation}).
Across 5,577 consecutive query pairs,\textbf{ 9\% are labeled as the same search}, \ie, the follow-up query differs only in word order, formatting, or synonyms (Figure \ref{fig:querysimilarity}).
For example, \gptoss follows \textit{``Turkey real GDP growth 2025''} with \textit{``2025 real GDP growth Turkey''}.
This behavior varies across models and datasets. \qwen repeats itself most often ($19\%$ on Products and $31\%$ on Nobel). 
The remaining queries do change the request, but often not in a targeted way.
Rather than addressing what the previous search failed to resolve, models frequently broaden the query with lists of candidate types or years, e.g., \claude follows \textit{`` `Desertion' release year''} with \textit{`` `Desertion' book film movie 2024 2023 2022 2021''}.

Repeated searches occur under both direct tool use and delegation ($7\%$ vs. $11\%$). The increase under delegation is mainly driven by \gpt ($8\%$ vs. $19\%$), while the other models behave similarly in both settings.

\xhdr{Orchestrators occasionally combine independent subtasks.}
We examine whether orchestrators delegate individual items as instructed or combine multiple items into a single executor request.
To identify such cases, we count how many items from the original question appear in each executor prompt.
Although items are independent and can be delegated separately, orchestrators sometimes combine them into a single request (Table \ref{tab:delegation-both-cutoffs}).
This behavior is most common for \gptoss ($4.5\%$ of calls under optional and $2.7\%$ under forced delegation) and is more frequent under optional delegation for \gpt ($3.2\%$ vs. $0.9\%$).
For example, one executor received \textit{``Provide the release year (four-digit) for each of the following book titles:'The Stone Sky', 'Lord of Shadows', 'Little Fires Everywhere', 'A List of Cages', and 'Dragon Teeth'...``}.
While rare, these cases occur within the same task distribution as correctly decomposed requests, following the same simple prompt setup that does not require joint reasoning.

Together, these findings highlight limitations in how models structure tool use, both when formulating searches within individual items and when decomposing tasks across items.

\begin{table}[t]
    \centering
    \small
    \setlength{\tabcolsep}{4pt}
    \begin{tabular}{@{}lcccc@{}}
        \toprule
        \textbf{Model}
        & \textbf{\gptoss}
        & \textbf{\qwen}
        & \textbf{\gpt}
        & \textbf{\claude} \\
        \midrule
        Single agent & 0.5 & 0.3 & 1.3 & 1.5 \\
        Forced       & 2.7 & 0.0 & 0.9 & 0.6 \\
        Optional     & 4.5 & 0.0 & 3.2 & 0.4 \\
        \bottomrule
    \end{tabular}
    \caption{\textit{Some models decompose queries into more than one item per executor.}
    Percentage of executor requests (single agent: web search queries) naming more than one item. Books and GDP datasets, both cutoff periods pooled.}
    \label{tab:delegation-both-cutoffs}
\end{table}

\subsection{Delegation Hinders Failure Recovery} %
Retrieval failures are common across tool-enabled agents.
In the single-agent setup, 13\% of \texttt{\small open\_url} calls return a 403 (Forbidden) error, while another 3\% return a page without extractable content.
These rates are higher under delegation: 20\% and 5\% for optional, and 17\% and 5\% for forced delegation.

\xhdr{Single agents remain more accurate after failed page visits.}
On items where a page visit returns a 403 error, single agents still answer approximately 73\% of affected items correctly, compared with 54\% under forced and 52\% under optional delegation.
A similar, smaller gap occurs for pages without extractable content: 60\% vs. 50\% vs. 45\%.
Thus, \textbf{single agents more often reach a correct answer despite encountering retrieval failures.}
These accuracies show that delegated setups are less successful on items where retrieval fails.

\xhdr{Communication boundaries can hinder error recovery.}
In the single-agent setup, retrieval errors remain visible to the agent. 
The agent producing the final answer directly observes retrieval errors, including which source failed.
It can use this information to change its strategy and pursue alternative searches or page visits.
Under delegation, only the executor's final response reaches the orchestrator, not its intermediate tool outputs.
The orchestrator may be unable to take corrective action, even when the executor has encountered an identifiable and potentially recoverable error. 
The orchestrator's ability to recognize an unresolved problem therefore depends on what the executor communicates.

Delegation therefore introduces an additional communication requirement: unresolved retrieval problems must be conveyed alongside the executor's findings.
This loss of error visibility may contribute to the observed lower accuracy of delegated setups.

\subsection{Orchestrators Challenge Executor Answers}

After receiving an executor's response, the orchestrator may question or reject its answer (\textit{challenging}) instead of using it for its own final answer. Alternatively, it may immediately delegate the same task again (\textit{re-delegation}). In both behaviors, the orchestrator can disregard information provided by an executor, even when that information is correct.
For example, on a Books query, the executor (\gptoss) correctly reported that \textit{A Gathering of Shadows} was published in 2016, but the orchestrator rejected this answer: \textit{``The subagent says 2016, but I think 'A Gathering of Shadows' was published in 2022 [...]. Subagent is wrong.''}

\begin{figure}
    \centering
    \includegraphics[width=\linewidth]{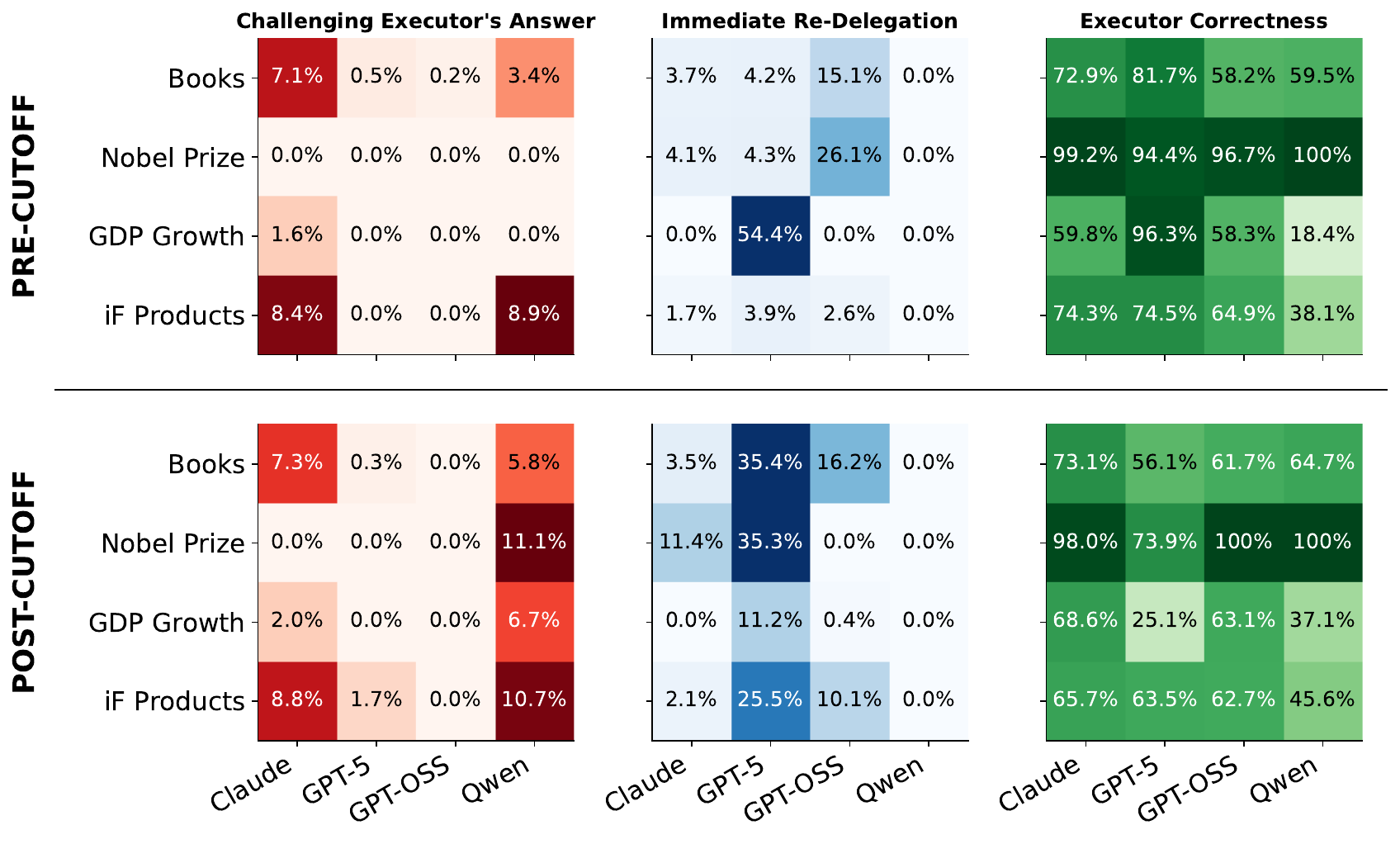}
    \caption{\textit{Models tend to either challenge executor responses or immediately re-delegate tasks. Re-delegation is more frequent post-cutoff, and both behaviors occur even when executors achieve high correctness.} \textbf{Left:} Share of challenged executor responses. \textbf{Middle:} Share of re-delegated executor responses of the same task. \textbf{Right:} Share of correct executor responses.
    }
    \label{fig:challenging_executor_responses}
\end{figure}

\xhdr{Models differ in how they respond to executor answers.}
Figure \ref{fig:challenging_executor_responses} showcases both behaviors across models, datasets, and cutoff conditions. While \claude and \qwen predominantly challenge executor responses, \gpt and \gptoss rarely challenge but frequently re-delegate tasks.
For example, \gpt re-delegates up to 54\% of tasks on pre-cutoff GDP. 
Across models and datasets, \textit{challenging} increases slightly from $2.6\%$ to $3.2\%$ (\uag{$23\%$}) between pre- and post-cutoff conditions, whereas \textit{re-delegation} increases more, from $6.6\%$ to $8.6\%$ (\uag{$30\%$}).
Thus, orchestrators seek additional executor responses more frequently for post-cutoff items.

\xhdr{Orchestrators re-delegate tasks even when executors are correct.}
To examine whether these behaviors are associated with incorrect executor responses, we condition on executor correctness.
Challenging is rare for correct executor responses, occurring in only $0.3\%$ of pre-cutoff and $0.2\%$ of post-cutoff cases.
In contrast, re-delegation remains common for correct executor answers, increasing from $6.4\%$ to $8.8\%$ (\uag{$39\%$}).
Orchestrators therefore request additional answers despite having the information needed to answer correctly, particularly post-cutoff.
This behavior happens even though the orchestrator is informed that the executor specializes in retrieving factual information.
While this verification may be useful, it introduces additional, potentially redundant, calls.

Together, these findings show that delegation can introduce additional work even after correct answer retrieval.
Producing correct information is therefore only one part of a successfully delegated task.
The orchestrator must also decide when that information is sufficient to complete the task.

\section{Can Prompting Mitigate Delegation Failures?} %
\label{sec:mitigations}
Recent work seeks to improve agent reliability through modifications to system prompts and agent harnesses \citep{lee2026meta,chen2026failed}.
We investigate whether similarly lightweight interventions can address the delegation failures identified in our analysis.
Specifically, we evaluate two prompt-based interventions targeting redundant searches and the communication of retrieval failures, and discuss why the remaining failure modes may not admit such simple solutions.

\xhdr{Discouraging redundant searches.}
To reduce inefficient tool use, we instruct agents to avoid semantically equivalent queries in successive web searches.
We evaluate this intervention on Books using \qwen and \gpt, which exhibit particularly high rates of redundant searches.
As shown in Figure~\ref{fig:addedbelief-norepeat} in Appendix~\ref{app:add-results}, the intervention has little effect on the semantic similarity of successive queries.
The share of query pairs judged as the same search goes down slightly for \qwen ($19.4\% \rightarrow 17.7\%$) while increasing slightly for \gpt ($4.4\% \rightarrow 6.6\%$).
The models also do not search less often overall, and the ranking between models remains unchanged.
Therefore, the instruction to avoid redundant searches neither cuts the number of follow-up searches nor makes them less redundant.

\xhdr{Reporting retrieval failures.}
To address the loss of information between executors and orchestrators, we instruct executors to explicitly report inaccessible webpages and their URLs.
We evaluate this intervention on GDP using \gpt and \gptoss, for which we observed high rates of webpage access failures.
For \gpt, accuracy on GDP increases from $0.43$ to $0.51$ with optional delegation and from $0.35$ to $0.44$ with forced delegation, but remains well below the single-agent baseline ($0.89$).
For \gptoss, accuracy on the GDP task drops from $0.70$ to $0.62$ for optional delegation and from $0.73$ to $0.64$ for forced delegation.
Explicitly requesting failure reports therefore offers limited, model-dependent improvements and does not reliably close the performance gap.

\xhdr{Why simple rules may not suffice.}
The remaining failure modes involve trade-offs that simple instructions may not resolve.
For instance, instructing models to formulate queries without assumptions could prevent adding misleading information.
However, prior knowledge can both help and hinder retrieval.
In an earlier example in Section ~\ref{sec:delegation-breaks}, the orchestrator incorrectly specified a work as a \textit{film}, steering the executor towards irrelevant results.
Conversely, specifying \textit{Kristen Ashley} as the author of \textit{Sebring} helped the executor identify the correct work (published in 2016).
Eliminating prior beliefs altogether may therefore impair retrieval effectiveness.

Similarly, simply instructing orchestrators to always accept executor responses could prevent them from incorrectly rejecting valid information.
In an earlier example, the orchestrator rejects the correct publication year (2016) returned by the executor in favor of its own incorrect belief (2022).
Yet executors can also return incorrect answers, for instance when multiple books with the same title exist in the search results.
The challenge is therefore not to eliminate all disagreement, but to distinguish reliable information from unsupported assumptions.
Addressing this challenge may require targeted mechanisms for communicating evidence and resolving conflicts rather than simple prompt interventions.

\section{Conclusion and Limitations}

Our work provides a systematic analysis of delegation for factual web search. By distributing independent tasks across separate contexts, delegation can reduce per-agent context requirements and mitigate the effects of context rot. 
However, our findings show that delegation does not consistently improve accuracy over a single agent with direct tool access.

Our analysis identifies several failure modes, many arising at the \textit{delegation boundary}: the separation of contexts between the orchestrator and the executors limits the information exchanged between them.
Critical information, such as retrieval failures faced by the executor, or supporting sources that could help the orchestrator build trust in the executor's response, may not reach the orchestrator.
The delegation boundary may therefore hinder recovery and make it harder for orchestrators to evaluate correct executor responses.
Instructing executors to report failures yields only limited improvements.
Passing complete executor traces to the orchestrator could mitigate the loss of information, but would undermine the context-saving benefits of delegation.
Other failures, such as redundant searches or injecting unsupported prior beliefs, are not inherent to delegation and may be easier to address.
However, prompting yields only limited and inconsistent improvements.
More effective mitigation may require targeted model training to induce specialization, tracking search progress, or mechanisms for selectively communicating evidence and failures. 
Whether such approaches generalize across tasks and models remains an open question.

\xhdr{Limitations.}
First, we focus on small datasets of independent factual questions. Real-world tasks may consist of interdependent information needs. 
We deliberately study this simpler setting to test delegation without the additional challenges of task interdependence. Failures in these simpler settings raise concerns about generalizability to more complex settings.
Second, we study configurations in which orchestrators and executors belong to the same model family. We do so to avoid a combinatorial explosion of heterogeneous setups. Studying the interplay between different model families is an interesting avenue for future work.

\bibliographystyle{plainnat}
\bibliography{custom}

\appendix

\appendix

\section{Experimental Details and Reproducibility}
\label{app:add-setup}

This appendix provides additional details on the datasets, models, agent configurations, execution setup, and tool call limits used in our experiments.

\xhdr{Models.}
\label{app:model_details}
We evaluate four models with tool-use capabilities, covering two closed-source and two open-weight model families. Within each experimental condition, the orchestrator and all executors are independent instances of the same underlying model. Table ~\ref{tab:models} reports the exact model identifiers and available knowledge cutoff information. When the provider reports a knowledge cutoff, we use that date. When no official cutoff is available, we instead report the model's release date. The release date provides a conservative upper bound on the model's possible training-time knowledge, since information published after the model itself was released could not have been included during its training. The actual knowledge cutoff may therefore be earlier than the reported release date.

\begin{table}[h]
    \centering
    \small
    \setlength{\tabcolsep}{4pt}
    \begin{tabularx}{\linewidth}{@{}cll>{\raggedright\arraybackslash}Xl@{}}
        \toprule
        & \textbf{Model}
        & \textbf{Provider}
        & \textbf{Identifier}
        & \textbf{Cutoff} \\
        \midrule

        \multirow{2}{*}{\rotatebox{90}{\shortstack{Closed\\source}}}
        & \claude
        & Anthropic
        & \texttt{claude-sonnet-4-5}
        & Jan 2025\footnotemark[1] \\

        & \gpt
        & OpenAI
        & \texttt{gpt-5}
        & Sep 30, 2024\footnotemark[2] \\

        \addlinespace[3pt]
        \midrule

        \multirow{2}{*}{\rotatebox{90}{\shortstack{Open\\weight}}}
        & \gptoss
        & OpenAI
        & \texttt{openai/gpt-oss-120b}
        & Jun 1, 2024\footnotemark[3] \\

        & \qwen
        & Qwen
        & \texttt{Qwen/Qwen3-30B-A3B-}\newline
          \texttt{Thinking-2507}
        & Jul 31, 2025\footnotemark[4]\textsuperscript{$\dagger$} \\

        \bottomrule
    \end{tabularx}

    \caption{
        Studied models and knowledge cutoffs. \textsuperscript{$\dagger$}For Qwen, the model release date is reported instead.
    }
    \label{tab:models}
\end{table}

\footnotetext[1]{\url{https://platform.claude.com/docs/en/models/sonnet-4-5/overview}}
\footnotetext[2]{\url{https://developers.openai.com/api/docs/models/gpt-5}}
\footnotetext[3]{\url{https://developers.openai.com/api/docs/models/gpt-oss-120b}}
\footnotetext[4]{\url{https://github.com/QwenLM/Qwen3}}

\xhdr{Generation Settings.}
\label{app:gen-settings}
The models were evaluated at a temperature of $T=1.0$, $top-p=1.0$. For the closed-source models, reasoning effort was set to medium and inference was performed through the respective provider APIs with their default sampling parameters. The open-source models were served with vLLM using the reasoning and tool calling parsers recommended in their HuggingFace documentation. Local inference was performed on NVIDIA H200 SXM GPUs. \qwen used vLLM's standard \texttt{\small /v1/chat/completions} endpoint, \gptoss used a client-side parsing fallback through \texttt{\small /v1/completions} to provide robust handling of malformed outputs observed with vLLM's built-in Harmony parser.

\begin{figure*}[h]
    \centering
    \includegraphics[width=0.9\linewidth]{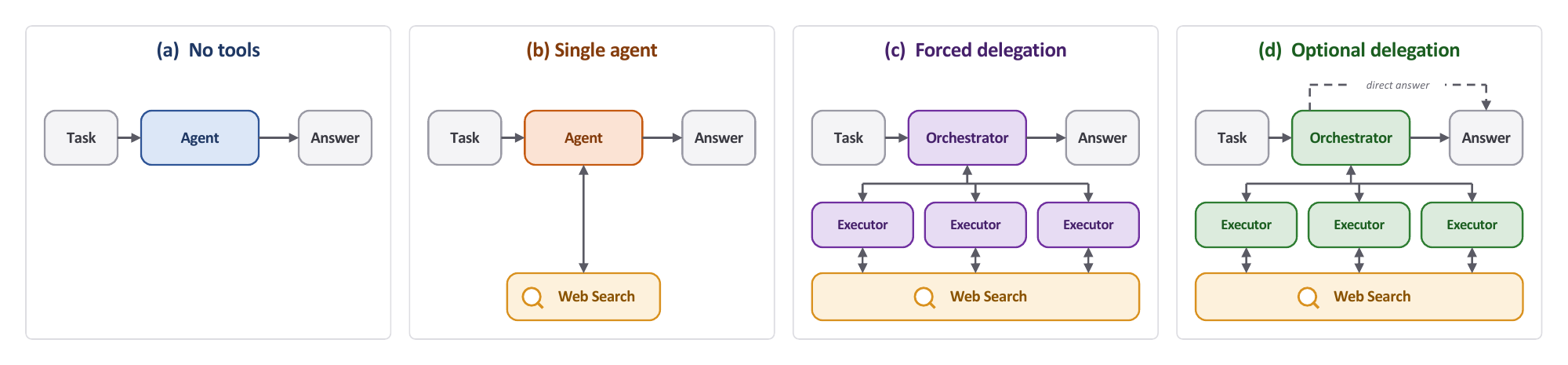}
    \caption{\textbf{Evaluated agent configurations.}
    (a) \textit{No tools}, (b) \textit{Single agent}, (c) \textit{Forced delegation}, (d) \textit{Optional delegation}.
    }
    \label{fig:agent-setups}
\end{figure*}

\xhdr{Agent Configurations.}
\label{app:agent_setup}
Figure~\ref{fig:agent-setups} illustrates the four evaluated agentic configurations. Table~\ref{tab:agent_prompts} reports the exact system instructions used in each condition.
In the \textit{no-tools} condition, a single model instance receives the complete user request and produces the final answer without access to external tools. In the \textit{single-agent} condition, the same model additionally has direct access to \texttt{web\_search} and \texttt{open\_url}. Retrieval and answer generation therefore take place within a single model trajectory.
In both delegation conditions, an orchestrator receives the complete user request but has no direct access to web search. Instead, it can invoke \texttt{call\_subagent} to send a self-contained subtask to an executor. Executors use the same underlying model as the orchestrator, but operate in separate contexts and have access to \texttt{web\_search} and \texttt{open\_url}. Each executor call starts with a fresh context containing the executor system instruction and the subtask formulated by the
orchestrator. After completing the subtask, the executor returns its final response to the orchestrator. The orchestrator receives neither the executor's intermediate trajectory nor the contents returned by its web tools.

Under \textit{forced delegation}, the orchestrator is instructed to delegate atomic subtasks rather than answer them itself. Under \textit{optional delegation}, the orchestrator may instead answer a subtask directly when it determines that its internal knowledge is sufficient. Executor calls are independent of one another, and executors do not observe the contexts or responses of other executors.

\xhdr{Datasets.}
\label{app:datasets}
Table~\ref{tab:dataset_overview} summarizes the data sources, temporal conditions, and number of prompts used in our experiments. Within each dataset and temporal condition, we use the same prompts across all models and agent configurations.

\begin{table}[h]
    \centering
    \footnotesize
    \setlength{\tabcolsep}{4pt}
    \begin{tabular}{lllcc}
        \toprule
        \textbf{Dataset} &
        \textbf{Source} &
        \shortstack[l]{\textbf{Years}\\\textbf{(Pre / Post)}} &
        \shortstack{\textbf{\# Prompts}\\\textbf{(Pre / Post)}} &
        \shortstack{\textbf{Items /}\\\textbf{Prompt}} \\
        \midrule
        Books    & Goodreads Top 100 \cite{goodreads_popular}
                 & 2014--2019 / 2026 & 120 / 20 & 5 \\
        Nobel    & Nobel Prize API \cite{nobelprize_api}
                 & 2014--2019 / 2025--2026 & 34 / 12 & 1--3 \\
        GDP      & World Bank API \cite{worldbank_api}
                 & 2019 / 2025 & 30 / 30 & 4 \\
        Products & iF Design Award \cite{ifdesign_awards}
                 & 2021 / 2026 & 30 / 30 & 4 \\
\bottomrule
    \end{tabular}
    \caption{Dataset sources and composition.}
    \label{tab:dataset_overview}
\end{table}

\xhdr{Tool Call Limits.}
\label{app:tool-call-limits}
To keep cost bounded and comparable across settings, we cap the number of tool calls per prompt. The limits scale with the number of items $n$ in a prompt. In the delegation settings, the orchestrator can call the executor at most $2n$ times. Accordingly, in the single-agent setup, the agent can call the web\_search function $2n$ times. For Nobel, $n$ is set to the maximum of three laureates per prompt. Table~\ref{tab:tool_call_limits} shows a detailed representation of all tool call limits across datasets and delegation setups.

\begin{table}[h]
    \centering
    \footnotesize
    \setlength{\tabcolsep}{4pt}
    \begin{tabular}{lcccc}
        \toprule
        & &
        \multicolumn{1}{c}{\textbf{Single Agent}} &
        \multicolumn{2}{c}{\textbf{Delegation (Forced / Optional)}} \\
        \cmidrule(lr){3-3} \cmidrule(lr){4-5}
        \textbf{Dataset} &
        \shortstack{\textbf{Budget}\\\textbf{Items ($n$)}} &
        \shortstack{\textbf{Search /}\\\textbf{Open URL}} &
        \shortstack{\textbf{Subagent}\\\textbf{Calls}} &
        \shortstack{\textbf{Search / Open URL}\\\textbf{per Subagent}} \\
        \midrule
        Books    & 5 & 10 / 15 & 10 & 2 / 3 \\
        Nobel    & 3 & 6 / 9   & 6  & 2 / 3 \\
        GDP      & 4 & 8 / 12  & 8  & 2 / 3 \\
        Products & 4 & 8 / 12  & 8  & 2 / 3 \\
        \bottomrule
    \end{tabular}
    \caption{Tool call limits per prompt.}
    \label{tab:tool_call_limits}
\end{table}

\begin{table*}[h]
\centering
\small
\renewcommand{\arraystretch}{1.2}
\setlength{\tabcolsep}{6pt}

\begin{tabularx}{\textwidth}{
    >{\raggedright\arraybackslash\bfseries}p{0.20\textwidth}
    >{\raggedright\arraybackslash}X
}
\toprule
\textbf{Condition / Role} & \textbf{System instruction} \\
\midrule
\rowcolor{gray!7}

No tools & You are a helpful assistant.
\\[0.5em]

\rowcolor{gray!7}
Single-agent & You are a helpful assistant with access to a \texttt{web\_search} and \texttt{open\_url} tool. If you do not know the answer to a question, use the \texttt{web\_search} tool. \\[0.5em]

\rowcolor{gray!7}
Forced-delegation orchestrator & You are a helpful assistant. Your task is to break the problem down into atomic subproblems and combine the answers into a final solution. Do not provide the answers to the subproblems
yourself, but use the tool \texttt{call\_subagent()} to delegate each subproblem to a subagent which will answer the subproblem for you. The subagent specializes in answering factual questions. Make sure each subproblem is self-contained. \\[0.5em]

\rowcolor{gray!7}
Optional-delegation orchestrator & 
You are a helpful assistant. Your task is to break the problem down into atomic subproblems and combine the answers into a final solution. If you know the answer to a subproblem, provide the answer yourself without calling the subagent. If you do not know the answer, use the tool \texttt{call\_subagent()} to delegate each subproblem to a subagent which will answer the subproblem for you. The subagent specializes in answering factual questions. Make sure each subproblem is self-contained. \\[0.5em]

\rowcolor{gray!7}
Executor & You are a helpful assistant with access to a \texttt{web\_search} and \texttt{open\_url} tool. Use them judiciously and only call a tool if you are not sure of the answer. If you know the answer and think the tool will not provide any additional information, do not use it. \\
\bottomrule
\end{tabularx}
\caption{Exact system instructions used in the different configurations.}
\label{tab:agent_prompts}
\end{table*}

\section{LLM Judge Prompts}
All LLM-based annotations used in our project are generated using \texttt{gpt-5.6-luna}. Judges are queried through the OpenAI Response API with medium reasoning effort and a maximum output length of 2,000 tokens.

\xhdr{Orchestrators Inject Incorrect Assumptions.}
To analyze how the orchestrator reformulates the user's request when delegating, we compare each delegated prompt with the corresponding information in the original user query. Each requested item is evaluated separately, even when a delegated prompt contains multiple items. For the single-agent setup we compare the first relevant \texttt{\small web\_search} query with the original user query. As an example, we show the judge prompt used for the Books dataset.
 
\begin{tcolorbox}[breakable, title=\textbf{\small Assumptions | \textit{Books}}]
\small
You are an evaluator comparing an original question with a rewritten question.

You will receive:
\begin{itemize}
    \item TARGET\_TITLE
    \item ORIGINAL\_QUESTION
    \item REWRITTEN\_QUESTION
\end{itemize}

TARGET\_TITLE is one of the titles in ORIGINAL\_QUESTION.
Only consider what REWRITTEN\_QUESTION says about TARGET\_TITLE.
If REWRITTEN\_QUESTION says something about several titles including
TARGET\_TITLE, it counts for TARGET\_TITLE.

The following are not changes:
\begin{itemize}
    \item Selecting TARGET\_TITLE from a list of titles.
    \item Turning the request into a standalone question or task.
    \item Rewording that keeps the meaning.
    \item Quotation marks, capitalization, or punctuation.
    \item Instructions about answer format or extra details to include in
    the answer (for example, ``return only the year'', ``specify the type'').
    \item General instructions about how to find the answer that do not name
    a specific source (for example, ``search the web'').
    \item Repeating what ORIGINAL\_QUESTION already asks for (for example,
    ``release year'').
\end{itemize}

For each field below, copy the exact words from REWRITTEN\_QUESTION that add
or change something about TARGET\_TITLE that is not stated in
ORIGINAL\_QUESTION and that narrows or changes which work is meant.
Judge only the wording. Do not use your own knowledge of the titles: an
addition counts even if it is correct (for example, ``novel'' for a title
that is a novel).
Count an addition even if it is a guess or one of several options (for
example, ``song or album'').
If nothing applies, return an empty list.

\begin{itemize}
    \item medium: the type of work (for example, book, novel, film, song).
    \item person: an author, director, actor, artist, or other person
    connected to the work.
    \item description: a genre, country, language, plot detail, or other
    description of the work (for example, ``American'', ``romantic drama'').
    \item time: a year, date, or period (for example, ``2014'', ``around 2015'').
    \item selection: a rule for choosing among works with this title (for
    example, ``the most notable one'', ``first published'').
    \item source: a specific source to use (for example, ``According to
    Wikipedia'').
    \item title: the title as written, if it differs from TARGET\_TITLE in
    more than capitalization or punctuation (for example, misspelled,
    shortened, or split).
    \item misc: anything else that is a change but does not fit the fields
    above.
\end{itemize}

Set reasoning to a 1--2 sentence explanation grounded in these definitions.
\end{tcolorbox}

\xhdr{Models repeatedly issue redundant queries.}
To assess repeated searches, the following judge prompt compares an initial web\_search query with a consecutive follow up query about the same item. For both delegation setups we analyze the two web\_search queries within one subagent call. For the single-agent setup we filter consecutive web\_search queries per item and compare these. The corresponding Figures \ref{fig:querysimilarity} and \ref{fig:addedbelief-norepeat} report the share of query pairs labeled as \textit{SAME}.
\begin{tcolorbox}[breakable, title=\textbf{\small Query similarity (SAME/DIFFERENT)}]
\small
You will see two web search queries. Decide whether they are the SAME search.\\

SAME: both queries contain exactly the same information. They may differ only in word order, formatting (quotes, capitalisation, punctuation) or by using a synonym (e.g. film/movie, novel/book).\\
DIFFERENT: one query adds, removes or changes any information, for example an author, a year, a source or website, the kind of item (book, film, song, ...), or words like "release year".\\

Examples:\\

< list of examples >\\

Now decide:\\
Query 1: <first query>\\
Query 2: <second query>\\
Answer with one word, SAME or DIFFERENT.
\end{tcolorbox}

\xhdr{Orchestrators Challenge Executor Responses.}
\label{app:orchestrator-challenges}
To identify cases in which information returned by an executor is challenged by the orchestrator, we annotate the orchestrator's reaction to each executor response. Executor answers returned within the same orchestration round are judged jointly, since the orchestrator may react to several answers in a single subsequent response. The judge assigns one of five labels to each executor answer: \textit{accepted}, \textit{questioned}, \textit{questioned then used}, \textit{rejected}, or \textit{no reaction}. 
We showcase the instruction for the books dataset as an example below.
For the analysis in Figure~\ref{fig:challenging_executor_responses}, we treat an executor answer as challenged when the judge label is \textit{questioned}, \textit{questioned then used}, or \textit{rejected}.

\begin{tcolorbox}[breakable, title=\textbf{\small Challenging | \textit{Books}}]
\small
You are checking how the orchestrator, an AI system that delegates
book lookups to a subagent executor, reacted to a list of subagent answers.
The goal of delegation is to elicit publication years of books.

You will receive:
\begin{itemize}
    \item TEXTS: a JSON list of the executor's answers, in the order they were
    received. Each answer's task is to state the publication year of a book.
    \item RESPONSE: what the orchestrator said right after receiving all of
    them. This may be empty.
\end{itemize}

For each item in TEXTS, decide how RESPONSE treated it.
Choose one reaction per item.

\begin{itemize}
    \item accepted: RESPONSE uses this text without expressing doubt.
    \item questioned: RESPONSE expresses doubt about this text. It does not
    give a different answer.
    \item questioned\_then\_used: RESPONSE expresses doubt about this text.
    However, it still uses the text's answer.
    \item rejected: RESPONSE gives a different answer than this text. Use this
    even if RESPONSE also expresses doubt.
    \item no\_reaction: RESPONSE does not address this item at all.
\end{itemize}

Return exactly one reaction for every item in TEXTS, in the same order.
Give a quote from RESPONSE that supports each reaction. A quote may be null
only for no\_reaction.
\end{tcolorbox}

\section{Additional Results}
\label{app:add-results}

\begin{figure}
    \centering
    \includegraphics[width=\linewidth]{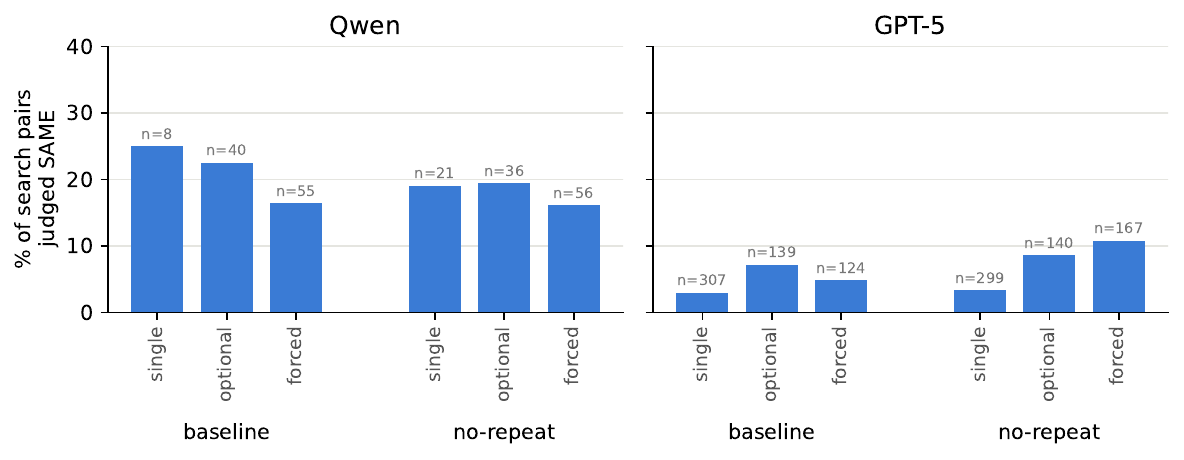}
    \caption{\textit{Explicitly instructing models to avoid repeating searches does not reduce redundancy.}
    The share of consecutive queries judged to be the same search remain essentially unchanged for both models on the Books dataset.}
    \label{fig:addedbelief-norepeat}
\end{figure}

\xhdr{Prompt-based Mitigations.}
We introduce a lightweight prompt-based mitigation to prevent models from repeatedly issuing semantically similar web search queries (see Section~\ref{sec:mitigations}).

\xhdr{Discouraging redundant searches.} We test whether explicitly instructing models to avoid semantically equivalent queries reduces redundant searches. Figure~\ref{fig:addedbelief-norepeat} compares the share of consecutive \texttt{\small web\_search} pairs classified as the same search under baseline and no-repeat instructions. The share of repeated pairs remains similar for both \qwen and \gpt. At the same time, models issued as many follow-up searches as before.

\begin{figure}
    \centering
    \includegraphics[width=\linewidth]{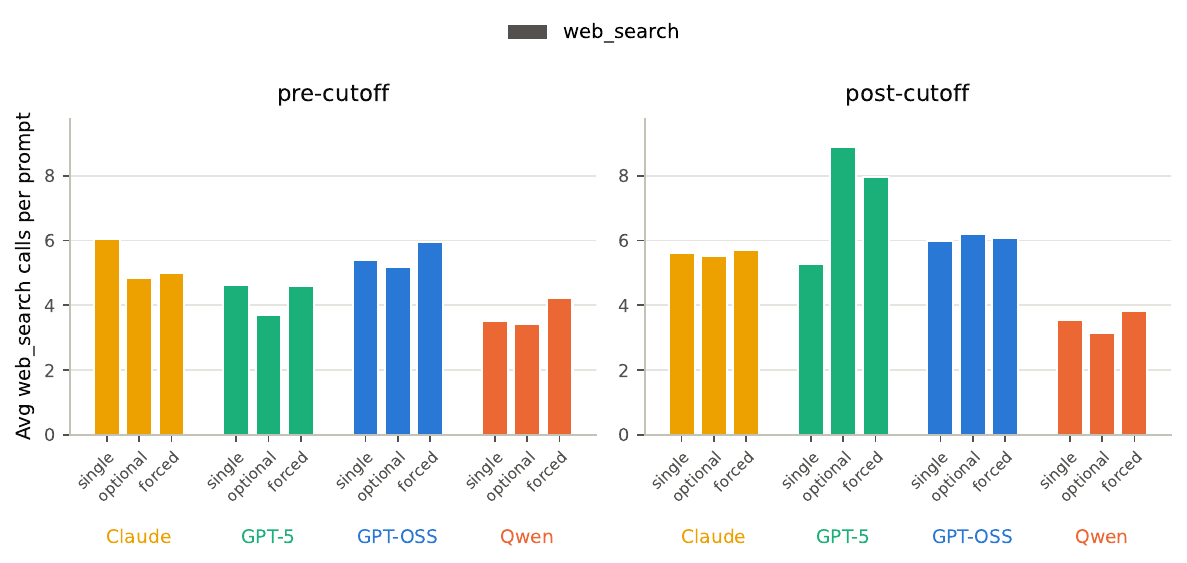}
    \caption{\gpt shows the largest increase in web search usage on post-cutoff questions. Average number of \texttt{\small web\_search} calls per question across datasets, shown by model and setting for pre-cutoff (left) and post-cutoff (right).}
    \label{fig:web_search_calls}
\end{figure}

\begin{figure}
    \centering
    \includegraphics[width=\linewidth]{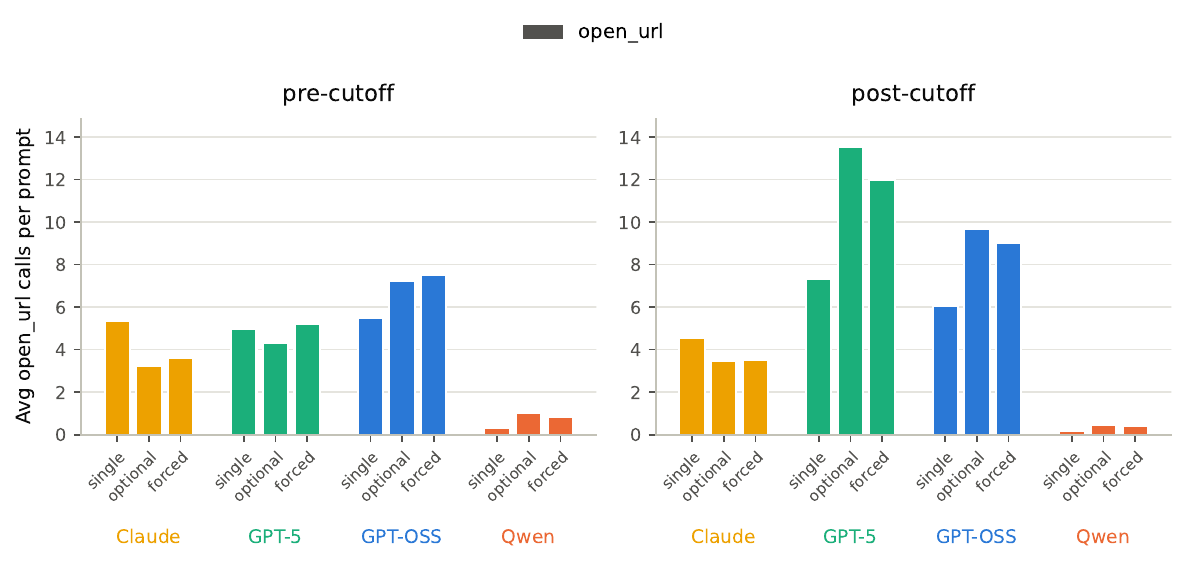}
    \caption{\gpt and \gptoss open more URLs on post-cutoff questions, while \qwen makes the fewest calls. Average number of \texttt{\small open\_url} calls per question across datasets, shown by model and setting for pre-cutoff (left) and post-cutoff (right)}
    \label{fig:open_url_calls}
\end{figure}

\xhdr{Number of Tool Calls.}
To quantify tool usage, we report the average number of \texttt{\small web\_search} (Figure~\ref{fig:web_search_calls}) and \texttt{\small open\_url} (Figure~\ref{fig:open_url_calls}) calls per prompt, averaged across the four datasets. We report results separately for pre-cutoff and post-cutoff questions under the single-agent, forced and optional delegation settings. 

Tool usage varies substantially across models and settings. \gpt  and \gptoss generally make more calls for post-cutoff questions, with the largest increases observed for \gpt in the forced and optional delegation settings. \claude exhibits relatively stable usage across the two periods, while \qwen consistently makes the fewest calls, particularly for \texttt{\small open\_url}.

\subsection{Human Validation of the LLM Judge}
\label{sec:human_validation}

To evaluate how closely the LLM judge (\texttt{gpt-5.6-luna}) agreed with human assessments, three authors independently annotated a random sample of 104 consecutive web-search query pairs from the Books dataset, using the same SAME/DIFFERENT definition given to the judge. 
The sample contains 13 pairs from each combination of orchestrator (\gptoss/\claude), cutoff period (pre/post), and setting (single-agent/forced delegation).
The annotators reached substantial agreement (Fleiss' $\kappa = 0.78$) and agreed on 101 of 104 pairs.
Two pairs whose first search failed are excluded from our analysis and were therefore not labeled by the judge.
On the remaining 102 pairs, the judge matched the annotators' majority vote in $97\%$ of cases ($\kappa =0.65)$, and its agreement with individual annotators ($\kappa = 0.65$--$0.85$) was in the same range as agreement among the annotators themselves ($\kappa = 0.65$--$0.90$). 
The three disagreements were borderline cases on which the annotators were also divided, such as ``film'' vs. ``short film''.

\end{document}